%% file: LuxRemix.tex
\documentclass[10pt,twocolumn,letterpaper]{article}

\usepackage{cvpr}              %

\input{preamble}

\definecolor{cvprblue}{rgb}{0.21,0.49,0.74}
\usepackage[pagebackref,breaklinks,colorlinks,allcolors=cvprblue]{hyperref}

\title{LuxRemix: Lighting Decomposition and Remixing for Indoor Scenes}

\author{
Ruofan Liang\textsuperscript{1,2} \qquad
Norman Müller\textsuperscript{1} \qquad
Ethan Weber\textsuperscript{1} \qquad
Duncan Zauss\textsuperscript{1} \\[0.25em]
Nandita Vijaykumar\textsuperscript{2,*} \quad
Peter Kontschieder\textsuperscript{1,*} \quad
Christian Richardt\textsuperscript{1,*}\\[0.5em]
\textsuperscript{1} Meta Reality Labs \quad
\textsuperscript{2} University of Toronto
}

\begin{document}

\twocolumn[{
\maketitle
\vspace{-1.2em}
    \centering
    \includegraphics[trim={0cm 0.4cm 0.6cm 0cm},clip,width=\textwidth]{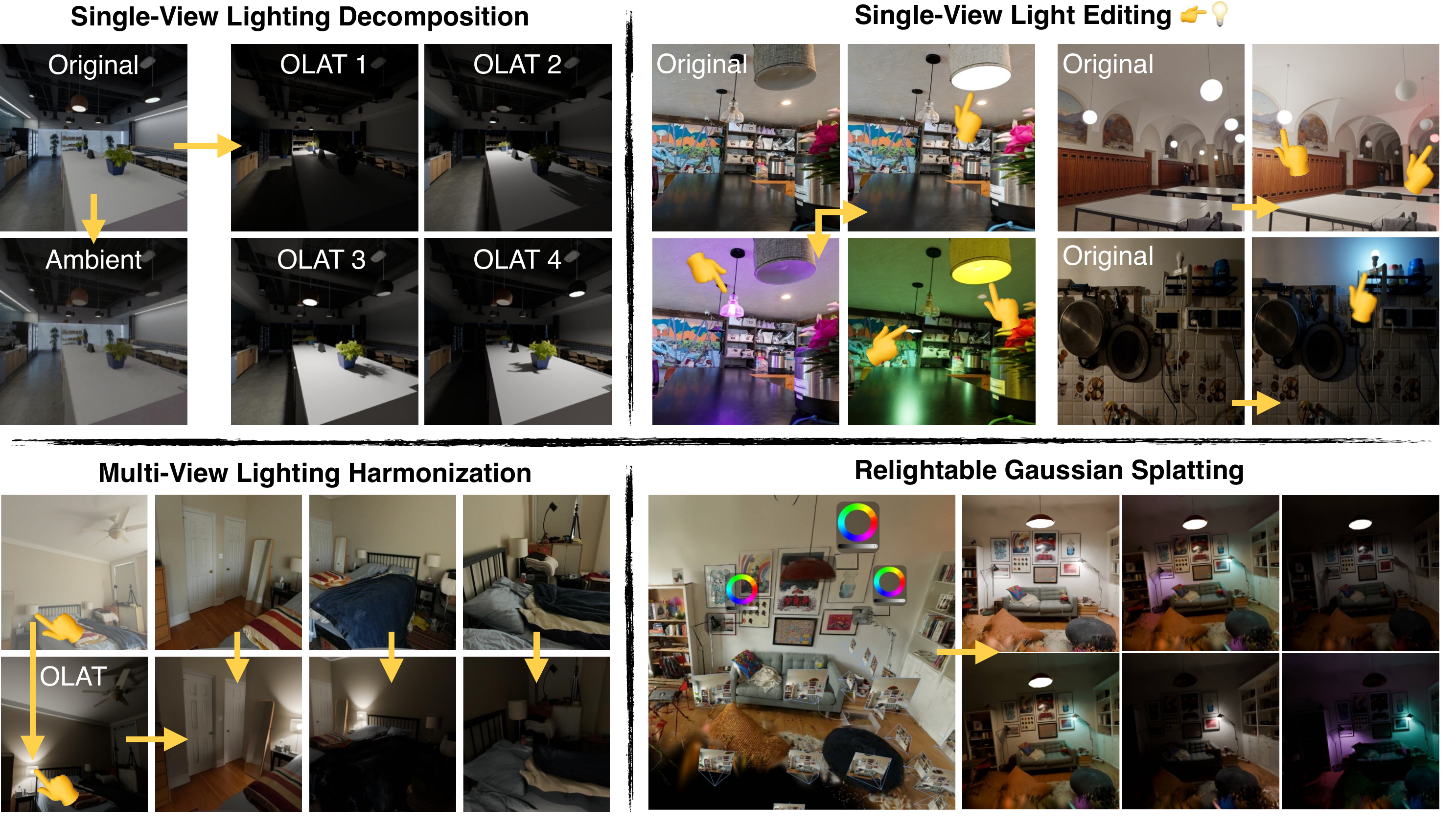}
    \captionsetup{hypcap=false}
    \captionof{figure}{%
        \paper enables interactive light editing of indoor scenes.
        Our method decomposes complex scene lighting into one-light-at-a-time (OLAT) sources and ambient lighting, which can be remixed for relighting effects.
        In the top right, we apply our method to single images, where we can change lights and their colors.
        \paper also enables multi-view-consistent harmonization of the decomposed lighting across multi-view images.
        By combining these capabilities, we enable real-time relighting of indoor scenes using 3D Gaussian splatting.
        \looseness-1
    }
    \label{fig:teaser}
    \vspace{0.2em}
}]

\iftoggle{cvprfinal}{%
{\let\thefootnote\relax\footnotetext{{* Joint advising}}}
}{}

\begin{abstract}
We present a novel approach for interactive light editing in indoor scenes from a single multi-view scene capture.
Our method leverages a generative image-based light decomposition model that factorizes complex indoor scene illumination into its constituent light sources.
This factorization enables independent manipulation of individual light sources, specifically allowing control over their state (on/off), chromaticity, and intensity.
We further introduce multi-view lighting harmonization to ensure consistent propagation of the lighting decomposition across all scene views.
This is integrated into a relightable 3D Gaussian splatting representation, providing real-time interactive control over the individual light sources.
Our results demonstrate highly photorealistic lighting decomposition and relighting outcomes across diverse indoor scenes.
We evaluate our method on both synthetic and real-world datasets and provide a quantitative and qualitative comparison to state-of-the-art techniques.
\iftoggle{cvprfinal}{%
For video results and interactive demos, see \href{https://luxremix.github.io/}{luxremix.github.io}.
}{}

\end{abstract}

\section{Introduction}
\label{sec:intro}

Controlling lighting in indoor scenes is fundamental to photography, cinematography, and virtual production workflows.
Professional photographers carefully adjust individual light sources to achieve desired aesthetic and functional lighting conditions.
However, this fine-grained control is typically lost after capture: images and 3D reconstructions bake lighting into the recorded appearance, making post-capture lighting adjustments difficult or impossible.
Enabling flexible post-capture editing of individual light sources would dramatically expand creative possibilities, allowing lighting decisions to be refined or alternative configurations explored without physical recapture.

Existing approaches to scene relighting face significant limitations.
Data-driven methods require dense multi-lighting captures under controlled conditions, which is impractical for real-world indoor scenes and offers limited generalization \cite{BiXSHHKR2020,BossJBLBL2021}.
Optimization-based inverse rendering can decompose scenes into geometry, materials, and lighting, but remains computationally intensive and often fails to produce plausible results when lighting conditions change significantly \cite{NimieDJK2021,WuZYZCMPLCR2023}.
Recent prior-driven approaches leverage pretrained diffusion models for relighting, but focus primarily on objects \cite{ZengDPKWT2024,JinLLXBZXSS2024}, portraits \cite{ChatuRHLDS2025}, or simple scenes under distant illumination \cite{LiangGLMHLGKVFW2025}.
Indoor scenes present unique challenges: spatially varying illumination from multiple near-field sources creates complex lighting interactions that are difficult to decompose and edit, especially when selectively switching individual lights on or off.
Moreover, existing single-image methods \cite{MagarHTPRSH2025} cannot maintain 3D consistency across multiple viewpoints, limiting their applicability to multi-view reconstructions.

We present \paper, a novel approach for interactive light editing in indoor scenes from a single multi-view capture.
Our method decomposes the complex lighting of an indoor scene into individually controllable light sources, enabling users to interactively adjust the intensity and color of each light or switch it on or off entirely.
The approach operates in three stages:
first, we leverage our proposed generative image-based light decomposition model to separate the contribution of each light source;
second, we propagate these decompositions consistently across all views using our multi-view lighting harmonization;
third, we train a relightable 3D Gaussian splatting representation that enables real-time interactive manipulation of individual light sources from any viewpoint.

Unlike existing methods that treat lighting as a global property or focus on distant illumination, our approach models spatially varying global illumination from multiple near-field sources in complex indoor environments.
While prior work on single-image lighting decomposition \cite{MagarHTPRSH2025} provides individual light control, it cannot ensure consistency across multiple views.
Conversely, multi-view relighting methods \cite{PhiliGZED2019,AlzayHBHSV2025} achieve 3D consistency but require controlled multi-lighting captures or produce only global relighting effects without per-light control.
Our method bridges this gap by combining the fine-grained control of single-image decomposition with the 3D consistency of multi-view methods.

The key insight enabling our approach is that modern diffusion models encode rich priors about indoor lighting that can be leveraged for decomposition, while multi-view geometric constraints provide the necessary consistency to propagate these decompositions across views.
By formulating lighting decomposition as a multi-view harmonization problem and encoding the results in a fast, differentiable 3D representation (3D Gaussians), we achieve both high-quality per-light control and real-time interactive performance.

Our main contributions are:
\begin{enumerate}
\item A single-image lighting decomposition model to factorize complex indoor scene lighting into individually controllable light sources.
\item A multi-view lighting harmonization method to ensure 3D consistency of the individual light source decompositions across all captured viewpoints.
\item Encoding the decomposed, consistent lighting in a relightable 3D Gaussian splatting representation, enabling real-time interactive manipulation of individual near-field light sources from novel viewpoints.
\item A large-scale synthetic dataset of over 12,000 generated indoor scenes with ground-truth per-light decompositions, which we are publicly releasing.
\end{enumerate}

\section{Related Work}
\label{sec:related}

There are three major approaches for editing scene lighting:
(1)~data-driven methods that require dense multi-lighting capture with limited generalization;
(2)~optimization-based inverse rendering that can be fragile for complex scenes;
and (3)~prior-driven methods that fine-tune pretrained models on task-specific data.
We adopt the prior-driven paradigm for its robustness and generalization.
This research direction has gained momentum over the last year, with most relighting work focusing on humans \cite{ZhangRA2025,ChatuRHLDS2025}, objects \cite{ZengDPKWT2024,JinLLXBZXSS2024}, or simple scenes \cite{LiangGLMHLGKVFW2025,LitmaTT2025} under distant illumination.
Scene relighting is more complex due to spatially varying illumination that is challenging to edit, e.g., to selectively switch lights on or off after the fact.

\subsection{Inverse Rendering and Image Decompositions}

Inverse rendering recovers scene geometry, materials, and lighting from images, enabling relighting and image editing.
Traditional optimization-based approaches \cite{NimieDJK2021,WuZYZCMPLCR2023,LinHLDRLZKWK2025} leverage differentiable rendering but are computationally intensive and may lack plausibility when lighting changes.
Recent neural and diffusion-based approaches have shown promising results, particularly for single images.

\paragraph{Single-image inverse rendering}

Early work on intrinsic decomposition \cite{GarceRCL2022,CareaA2023} separated images into shading and albedo components, which was later extended by diffusion-based approaches \cite{LuoCYZPFLRW2024,ZengDGHHLYH2024,SunWZXRFXY2025} that jointly estimate intrinsic layers and enable material-aware synthesis.
For complex indoor scenes, methods have progressively improved from recovering shape, lighting, and SVBRDF \cite{LiSRSC2020,LiSBZSHXRC2022,MunkbHSGCEMF2022} to using vision transformers \cite{ZhuLMPC2022} and diffusion models \cite{KocsiSN2024,LiangGLMHLGKVFW2025} for more robust decomposition.

\paragraph{Multi-view inverse rendering}

Multi-view methods can reconstruct more accurate geometry and appearance than single-view inverse rendering, but require multi-view input.
Early approaches learned volumetric reflectance representations \cite{BiXSHHKR2020} and pre-integrated lighting networks \cite{BossJBLBL2021,ChoiKK2023a}, while differentiable rendering enabled extraction of meshes with physically-based materials \cite{NimieDJK2021,MunkbHSGCEMF2022,HasseHM2022,WuZYZCMPLCR2023}.
Recent work has expanded to unconstrained captures \cite{BossEKLSBLJ2022,EngelRBZKLSBBLJ2024,LiWCLNLHXZZMRND2025}, large-scale scenes \cite{LiWCPY2023,ZhuHYLLXWTHBW2023}, and practical scenarios including near- and far-field sources \cite{FeiTTS2024}, low dynamic range inputs \cite{LinHLDRLZKWK2025}, and physics-based indirect illumination \cite{DengLLY2024,ChoiLPJKC2025}.

\paragraph{Single-view lighting decomposition and editing}

Lighting decomposition and editing enables flexible workflows where lighting is manipulated after capture \cite{BoyadPB2013,MekaSZRT2021}.
\citet{EinabGH2021} provide a comprehensive survey of deep neural models for illumination estimation and relighting.
While latent-based methods \cite{BhattSF2024,XingGKGB2025} achieve complex relighting effects, they lack detailed control over individual light sources.
LightLab provides parametric control over individual light source intensity and color by fine-tuning diffusion models on photograph pairs \cite{MagarHTPRSH2025}.
Our aim is to extend this capability to multi-view captured scenes.

\begin{figure*}[t]
    \centering
    \includegraphics[width=\textwidth,trim={1.5cm 1cm 2.9cm 0cm},clip]{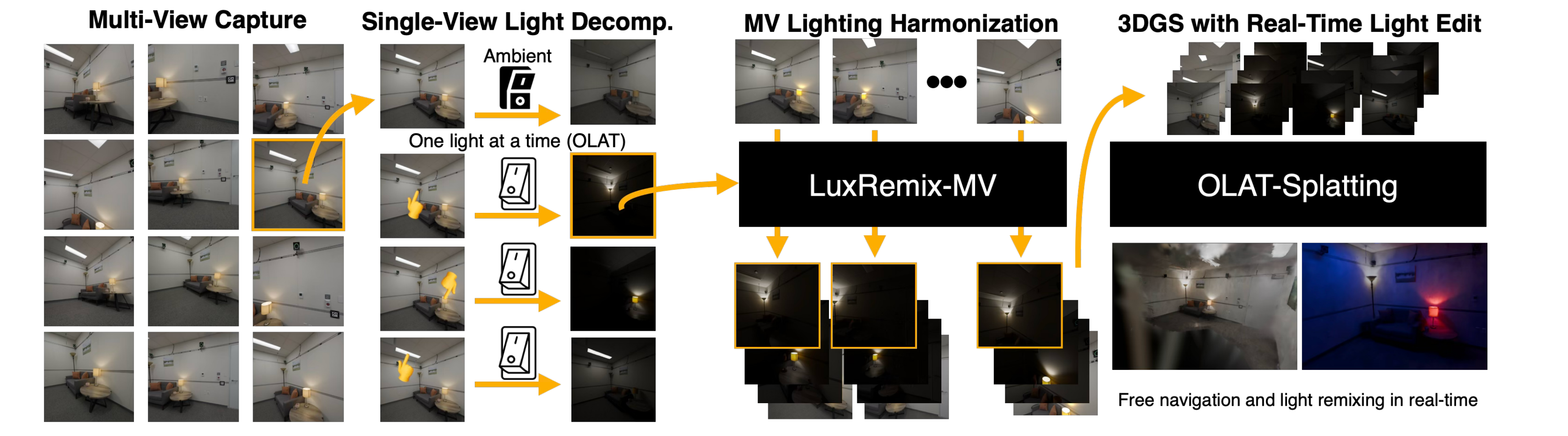}
    \caption{\textbf{Overview.}
    From a multi-view capture, our single-image model decomposes each light source into a one-light-at-a-time (OLAT) image, which we propagate across views via multi-view harmonization with geometric constraints.
    The resulting multi-view OLAT dataset is used to train a relightable 3D Gaussian splatting representation for real-time interactive light control from any viewpoint.
    }
    \label{fig:pipeline}
\end{figure*}

\subsection{Image- and Video-based Relighting}

Prior-driven relighting approaches involve training or fine-tuning a pretrained model on large datasets with \mbox{(pseudo-)ground-truth} relighting pairs.
This allows the model to learn lighting priors and appearance changes under varying illumination, often generalizing well to unseen scenes, especially when fine-tuned on real-world data.

\paragraph{Single-image relighting}

Recent diffusion-based methods enable detailed lighting control for object relighting, including DiLightNet \cite{ZengDPKWT2024}, Neural Gaffer \cite{JinLLXBZXSS2024}, and IC-Light \cite{ZhangRA2025}.
For portrait relighting with environment maps, methods range from light stage training \cite{PandeOLHBRDF2021,ZhangZWYX2021} and 3D-aware representations \cite{MeiZZSZBZJP2024,RaoFMMZWBPMET2024} to physics-driven architectures \cite{KimJYLNW2024} and diffusion models trained on synthetic faces \cite{ChatuRHLDS2025}.
However, environment maps assume distant illumination, which is unrealistic for indoor scenes.
Relighting without environment maps can leverage the background image as a reference \cite{RenXYSZJGZ2024,WangLSSSZYZWCNY2025}.
For outdoor scenes, methods range from self-supervised inverse rendering \cite{YuMESTS2020} to shadow prediction via learned ray-marching \cite{GriffRP2022} or conditional diffusion models \cite{KocsiPSNH2024}.
For indoor scenes, diffusion-based methods enable explicit control via light source parameters \cite{MagarHTPRSH2025} or user scribbles \cite{ChoiWPBS2025}, while \citet{CareaA2025} combine path-tracing with neural rendering for physically-based relighting.

\paragraph{Video relighting}

Extending image relighting to video introduces temporal consistency challenges, as frame-by-frame application produces lighting and appearance flicker.
Training-free approaches \cite{ZhouBLZWHLDZCRWN2025} adapt image models \cite{ZhangRA2025} with cross-frame attention, while end-to-end methods \cite{ZengLFMGQZWY2025,MeiHMPXGYDTYPD2025} leverage video diffusion models trained on hybrid synthetic-real datasets.
To avoid error accumulation in two-stage pipelines, DiffusionRenderer \cite{LiangGLMHLGKVFW2025} uses G-buffers to decouple inverse and forward rendering, while UniRelight \cite{HeLMHVKFGGW2025} jointly estimates albedo and synthesizes outputs in one pass.

\paragraph{Multi-view relighting}

Multi-view relighting methods leverage geometric information to achieve 3D consistency and novel view synthesis.
Early methods used proxy geometry from multi-view stereo with geometry-aware networks \cite{DucheRCLLPBD2015,PhiliGZED2019}.
NeRF-in-the-Wild \cite{MartiRSBDD2021} and GaRe \cite{BaiZJHLLGFGC2025} use appearance embeddings for varying illumination, which lacks explicit relighting control.
Recent diffusion-based approaches synthesize multi-illumination data from single-illumination captures \cite{PoiriGPLD2024} or relight input images before reconstruction \cite{ZhaoSVPBH2024,AlzayHBHSV2025,LitmaTT2025}, avoiding brittle inverse rendering optimization.

\subsection{3D-Consistent and Real-time Relighting}

\paragraph{Gaussian splatting for inverse rendering}

3D Gaussian splatting \cite{KerblKLD2023} has been extended to inverse rendering, addressing challenges in normal estimation and occlusion via regularization \cite{LiangZFSJ2024}, ray tracing \cite{GaoGLZCZY2024}, or hybrid representations \cite{YeGLCC2025}.
Methods also tackle global illumination \cite{ChenLZ2025}, spatially-varying HDR lighting \cite{BolduHSL2025}, diffusion-guided material decomposition \cite{DuLSW2025}, and specialized effects including inter-reflections \cite{GuWZYZ2025,LiangLJGZ2025} and radiance transfer \cite{ShiWWLZFZZDW2025,ZhouZWZ2025,LiuGLBZ2025}, for high-quality relighting results.

\paragraph{Multi-view harmonization and consistency}

Multi-view harmonization reconciles inconsistent illumination across captured views for accurate 3D reconstruction.
\citet{AlzayHBHSV2025} use diffusion models to harmonize inputs to a reference illumination, while LightSwitch \cite{LitmaTT2025} incorporates material guidance for consistent relighting.
CAT3D \cite{GaoHHBMSBP2024} and SEVA \cite{ZhouGVVYBTRJ2025} generate consistent novel views, while SimVS \cite{TreviPHVWAGPBHRS2025} trains on synthetic data that simulates illumination variation.

\paragraph{Real-time relightable representations}

Neural rendering approaches combine multi-view stereo geometry with learned material and illumination representations for interactive relighting \cite{PhiliMGD2021}.
3D Gaussians have recently emerged as a powerful representation for real-time relighting, including precomputed radiance transfer \cite{ZhangHLLX2024}, BRDF decomposition \cite{GaoGLZCZY2024,BiZZPFZW2024,JiangTLGLWM2024}, diffusion-guided estimation \cite{DuLSW2025}, neural features \cite{FanLYHW2025,LiuGLBZ2025}, and multi-bounce lighting \cite{HuGWL2025}, as well as domain-specific methods for avatars \cite{SaitoSSLN2024,WangSSBLAPYNGLWRSZGTS2025,SchmiGN2025} and outdoor scenes \cite{BaiZJHLLGFGC2025}.
However, most methods require controlled multi-view OLAT capture rather than casual captures.

\section{Multi-view Light Editing}
\label{sec:method}

Our goal is to decompose the complex lighting of an indoor scene into individually controllable light sources given a single multi-view capture.
This enables users to interactively adjust the intensity and color of each light or switch it on or off, with view-consistent results.
\cref{fig:pipeline} provides an overview of our three-stage approach.
First, we train our single-image lighting decomposition model for per-view generative light decomposition (\cref{sec:single_image}).
To achieve this, we generate a large-scale synthetic dataset with ground-truth per-light decompositions (\cref{sec:synthetic_data}), enabling us to fine-tune pretrained diffusion models that encode rich priors about indoor lighting.
Second, we propagate these decompositions consistently across all views using multi-view geometric constraints through a lighting harmonization framework (\cref{sec:lighting_harmonization}).
Third, we encode the decomposed lighting in a relightable 3D Gaussian splatting representation that enables real-time rendering and interactive manipulation of individual light sources from any viewpoint (\cref{sec:relightable_3dgs}).

\input{assets/synthetic_data_example.tex}

\input{assets/single_image_light_editing.tex}

\subsection{Synthetic Multi-Light Data Generation}
\label{sec:synthetic_data}

A key component of our method is the synthetic multi-light dataset used to train our single-image lighting decomposition and multi-view harmonization models.
We start with over 12,000 procedurally generated 3D models of indoor scenes \cite{AvetiXHYAPZFHOEMNB2024}, augmenting each scene with procedurally generated light sources using Infinigen \cite{RaistMKYZHWPALMD2024} with up to six controllable lights in total.
This includes ceiling, wall, floor, and table lamps, as well as environment lighting \cite{KlotzN2025,PolyHaven2025}.
The emitted light source colors are sampled from black-body color temperatures to cover a spectrum of white light and augmented with 10\% HSV variations to cover a wider range of colors.

We render each scene using Blender's Cycles renderer \cite{BOC2025} with all lights on and in multiple one-light-at-a-time (OLAT) configurations, where all lights except one are switched off.
To enable flexible viewpoint sampling during training, we render each scene as four equirectangular HDR images at roughly eye height rather than pre-rendering all possible perspective views, allowing us to sample diverse camera views on the fly while avoiding the high upfront rendering cost.
For each light source, we generate three mask types: (1)~the emissive area (dilated if too small), (2)~the full light fixture, and (3)~its convex hull.
For further details, please see the supplement.
\Cref{fig:synthetic_data_example} shows examples of our synthetic multi-light data with various light configurations.

\subsection{Single-image Lighting Decomposition}
\label{sec:single_image}

Given a single input image of an indoor scene, $I_\text{input}$, we decompose the lighting into ambient lighting, $I_\text{ambient}$, and multiple one-light-at-a-time (OLAT) light sources, $I_i$, that can be edited individually:
\begin{align}
    I_\text{input} = \text{tonemap}\Big(I_\text{ambient} + \sum_{i=1}^{N} \boldsymbol{c}_i \cdot I_i\Big) \text{.} \label{eq:image_formation}
\end{align}
The OLAT images are determined up to scale, requiring RGB scale factors $\boldsymbol{c}_i$ to recreate the input lighting.

We fine-tune a pretrained image editing DiT model using LoRA \cite{HuSWALWWC2022} to enable flexible light editing given the additional spatial prompt of the light selection mask, to decompose the input image into $I_\text{ambient}$ and OLAT passes $I_i$.
Our LoRA fine-tuning enables the DiT to focus on two light editing tasks:
1) OLAT decomposition using text instructions like “switch off all lights except the selected one” to generate an edited image with isolated individual light contribution; and
2) Turning off the light by prompting the model to “switch off only the selected light” to show the scene illuminated by all remaining sources, which we treat as ambient lighting.
To condition on which light to edit, we patchify the light mask, tokenize it via a single-layer MLP, and add these tokens channel-wise to the input image features.
We augment training data via light composition, dynamically combining multiple OLAT images.
We also prompt with three brightness levels (high/medium/low, matching EV0/EV-2/EV-4 of the target HDR OLAT) to improve light transport learning and capture a wider dynamic range.
\Cref{fig:single_image_light_editing} shows single-image lighting decomposition results.

\input{assets/lighting_harmonization.tex}

\subsection{Multi-view Lighting Harmonization}
\label{sec:lighting_harmonization}

While existing methods can decompose lighting in single images or generate consistent multi-view imagery from scratch, no prior work addresses the challenge of propagating lighting decompositions across multi-view captures.
This novel task requires both understanding the 3D scene geometry and maintaining photometric consistency while transferring complex lighting from conditional views to input views.

Given multi-view input images where lighting has been decomposed in one or more views,
we aim to propagate the lighting consistently across all remaining views.
Our approach takes as input the multi-view images with partial lighting decompositions along with their corresponding Plücker ray embeddings, and produces multi-view images with harmonized lighting across all views.
Inspired by recent multi-view diffusion methods like CAT3D \cite{GaoHHBMSBP2024}, SimVS \cite{TreviPHVWAGPBHRS2025}, and SEVA \cite{ZhouGVVYBTRJ2025},
we follow the latter's approach to propagate lighting consistently while preserving geometric consistency.
Similar to SimVS, we concatenate the original input views and sparse light-decomposed views along with the corresponding Plücker ray embeddings and reference view masks to the pretrained multi-view diffusion U-Net, and perform full-parameter fine-tuning of the U-Net.
Please refer to the supplement for more details.

To generate high dynamic range (HDR) outputs, we run the lighting harmonization process three times with different exposure levels for each lighting condition (“high, medium, low” from \cref{sec:single_image}).
For each view and lighting condition, we then merge the three exposure-bracketed images to create the final HDR output following \citet{DebevM1997}. 
This per-light HDR reconstruction allows us to maintain the full dynamic range of the scene while ensuring consistent lighting across all viewpoints.
\Cref{fig:lighting_harmonization} shows our multi-view lighting harmonization results.

\subsection{Real-time Remixable Lighting in 3D}
\label{sec:relightable_3dgs}

\begin{figure}[t]
\includegraphics[width=\linewidth]{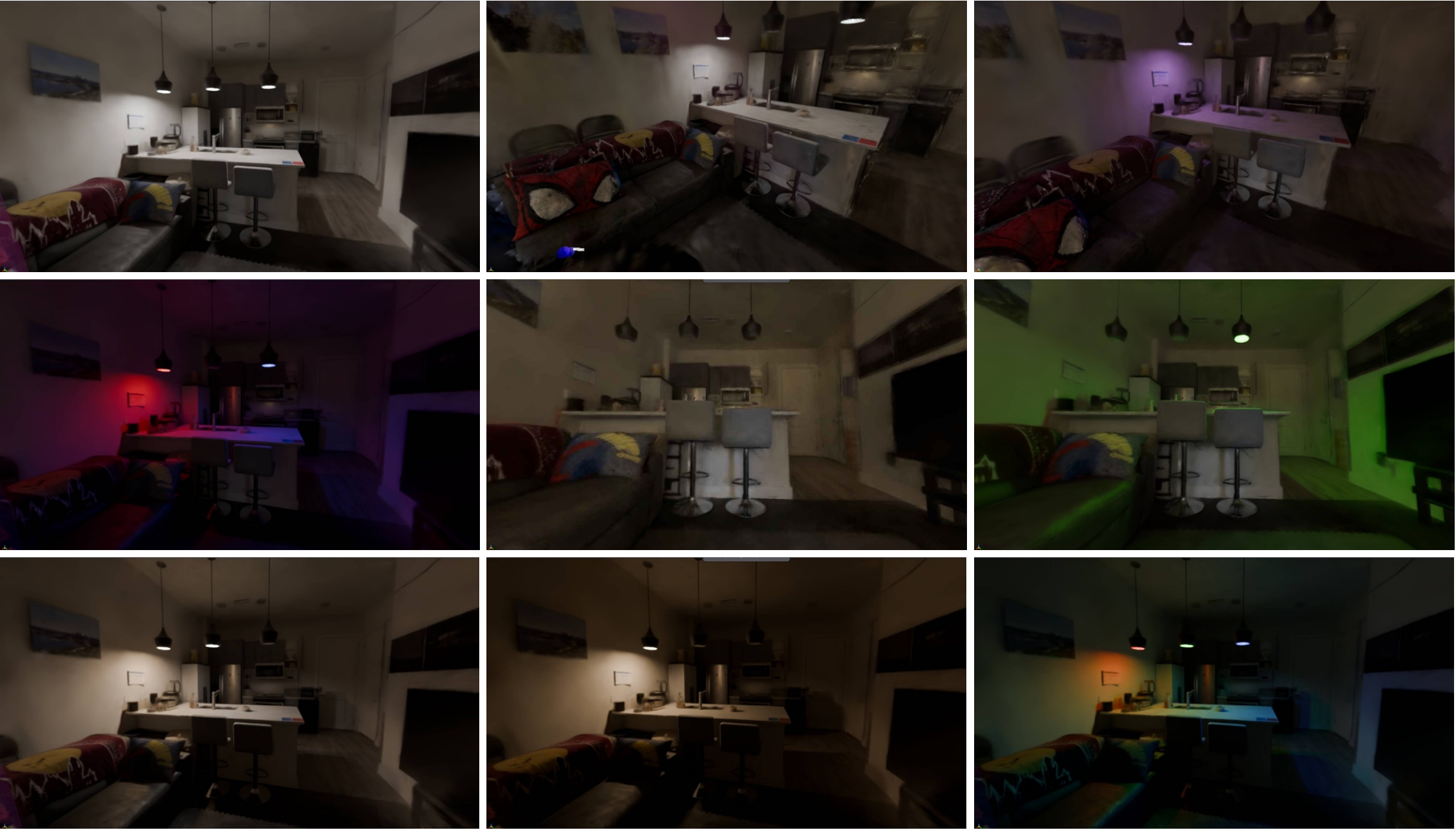}
\caption{\label{fig:relightable_3dgs}%
    \textbf{Real-time Remixable Lighting.}
    Original lighting (top left) and interactively created lighting under varying viewpoints.
} 
\vspace{-0.5em}
\end{figure}

We encode the decomposed lighting into a relightable 3D Gaussian splatting representation that enables real-time rendering and interactive manipulation of individual light sources from any viewpoint (see \cref{fig:relightable_3dgs}).
Building upon 3D Gaussian splatting \cite{KerblKLD2023}, we extend each Gaussian with per-light HDR RGB coefficients, storing the appearance contribution from each light source (including ambient lighting) separately.
At render time, we linearly combine these per-light contributions with user-controlled light intensity or color to produce the final appearance under arbitrary lighting configurations.
This representation preserves the real-time rendering capabilities of standard 3D Gaussian splatting while enabling independent control over each light source.

We optimize this representation in two stages.
First, we pretrain a standard 3D Gaussian splatting model \cite{YeLKTYPSYHTK2025} on the original multi-view input images to establish the geometric structure and spatial distribution of Gaussians.
Second, we freeze all geometric and appearance parameters, then introduce per-light RGB coefficients for each Gaussian, initialized from the multi-view harmonization outputs, and optimize them using the decomposed multi-view lighting images.
We train these coefficients in linear HDR space with differentiable tone mapping, jointly solving for per-Gaussian and per-light color scaling factors that ensure the recombined lighting matches the original input images.
Our training uses an L1 loss for the individual per-light images and an L1 composition loss for consistency with the original input views.

\section{Evaluation}
\label{sec:evaluation}

To quantitatively evaluate our single-view lighting decomposition and multi-view lighting harmonization models, we use 30 synthetic test scenes held out from training.
We evaluate our models on this test set using PSNR, SSIM, and LPIPS \cite{ZhangIESW2018} after channel-wise color rescaling w.r.t. the ground truth.
For real-world scenes, we use standard SfM to estimate camera poses and typically use 32--96 images covering sufficient views of the target lights.

\subsection{Single-Image Lighting Decomposition}

We compare our \paper-SV model for single-image lighting decomposition against two baseline methods:
ScribbleLight \cite{ChoiWPBS2025}, a diffusion-based approach for scribble-guided lighting editing, and
Qwen-Image \cite{WuLZLGYYBXCCTZWYYCLLZMWNCCPQWWYWFXWZZWCL2025}, a general-purpose image editing foundation model.
We also evaluate several ablations of our model.
The `FLUX token' variant is fine-tuned as an in-context LoRA \cite{HuangWWSDLFLZ2024} by concatenating the input image and mask side-by-side as tokens.
The `SD' variant uses a U-Net-based latent diffusion model \cite{RombaBLEO2022}.
Our final \paper-SV model uses channel-wise token addition where the mask is processed by a single-layer MLP to match the FLUX VAE latent dimensions, then added to the input condition image latents.
Similar to the `FLUX token' variant, we use LoRA for parameter-efficient fine-tuning.

\begin{figure}
\includegraphics[width=\linewidth]{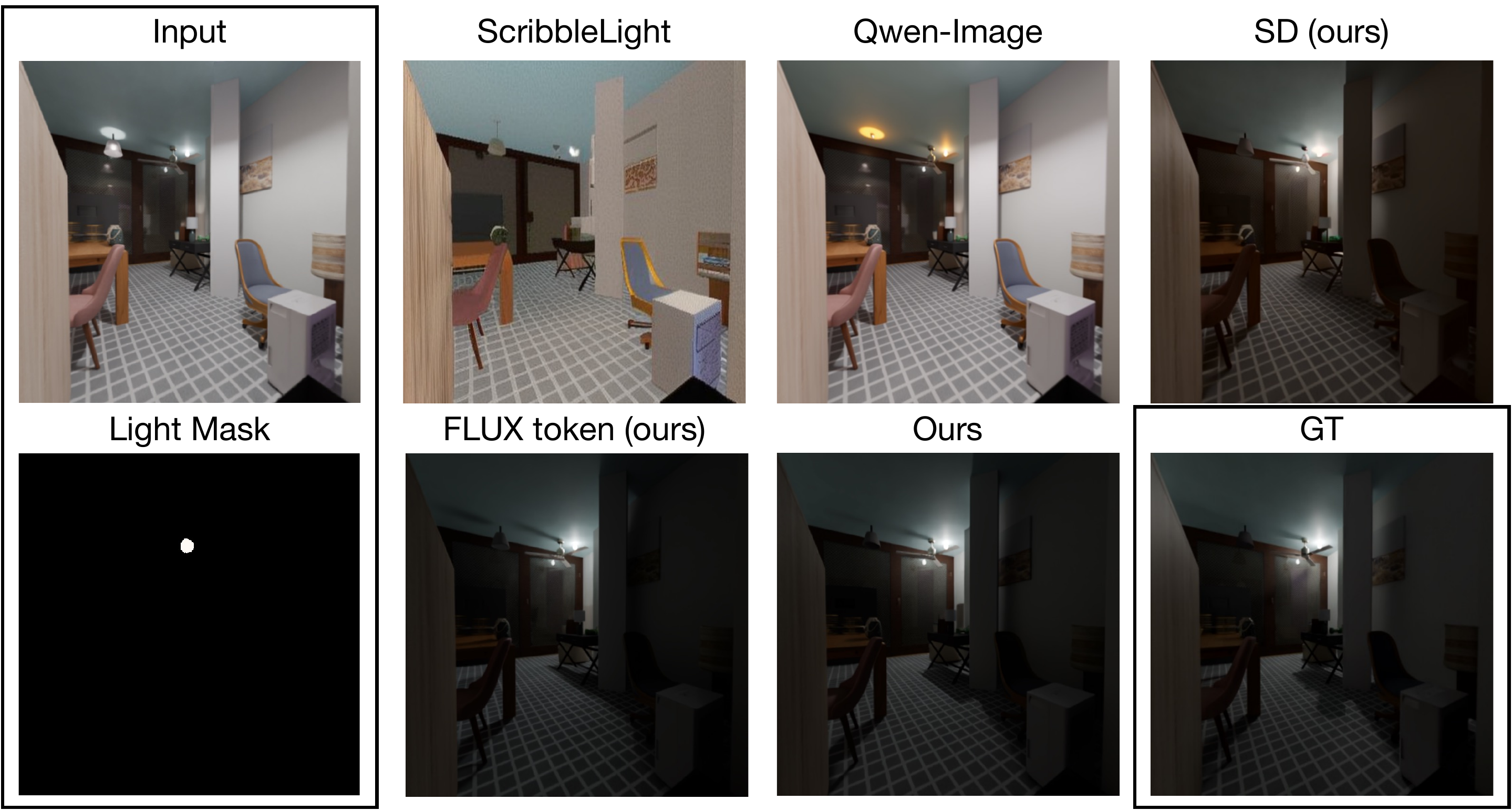}
\caption{\label{fig:single_image_qualitative}%
    \textbf{Qualitative Single-Image Lighting Decomposition.}
    Given an input image and light mask, we compare methods for isolating individual light sources (ground truth at bottom right).
    Our method most accurately decomposes the lighting.
}
\end{figure}

\Cref{tab:single_image_quantitative} and \cref{fig:single_image_qualitative} show results for our single-image lighting decomposition model, which successfully isolates individual light contributions by switching off all lights except the selected one.
Existing image editing models lack precise per-light control capability while our final method achieves the best performance across all metrics.
\Cref{fig:single_image_comparison} shows a qualitative comparison of our \paper-SV model with the baseline methods for light editing.

\subsection{Multi-View Editing/Harmonization}
\Cref{fig:lighting_harmonization} shows qualitative results
of our multi-view lighting harmonization model, which successfully propagates lighting consistently across all views.
We also quantitatively compare our \paper-MV model against two ablations in \cref{tab:multi_view_harmonization_quantitative}:
1) \paper-SV processes each view independently without multi-view context, and
2) \paper-MV-Edit, which extends \paper-MV with additional light masks to perform mask-guided multi-view editing, instead of lighting harmonization from sparse reference views.
The metrics demonstrate the necessity of our \paper-MV model.

\begin{table}
\centering
\setlength\tabcolsep{4pt}
\caption{\textbf{Single-Image Lighting Decomposition.}
Our method and variants compared to baselines on 30 synthetic test scenes.}
\label{tab:single_image_quantitative}
\resizebox{0.85\linewidth}{!}{
\begin{tabular}{l
    S[table-format=2.2, round-mode=places, round-precision=2, detect-weight=true, mode=text]
    S[table-format=1.3, round-mode=places, round-precision=3, detect-weight=true, mode=text]
    S[table-format=1.3, round-mode=places, round-precision=3, detect-weight=true, mode=text]}
\toprule
\bf Method & {\bf PSNR $\uparrow$} & {\bf SSIM $\uparrow$} & {\bf LPIPS $\downarrow$} \\
\midrule
ScribbleLight \cite{ChoiWPBS2025} & 14.393476 & 0.395403 & 0.687826 \\
Qwen-Image \cite{WuLZLGYYBXCCTZWYYCLLZMWNCCPQWWYWFXWZZWCL2025} & 18.233805 & 0.713984 & 0.237084 \\
\midrule
\multicolumn{4}{l}{\textit{Our variants}} \\
~~FLUX token \cite{FLUX2025} & 25.198421 & 0.864664 & 0.101492 \\
~~SD \cite{RombaBLEO2022} & 27.134128 & 0.857183 & 0.099066 \\
\midrule
\paper-SV & \bfseries 27.676194 & \bfseries 0.898317 & \bfseries 0.081680 \\
\bottomrule
\end{tabular}
}
\end{table}

\begin{table}
\centering
\setlength\tabcolsep{4pt}
\caption{\label{tab:multi_view_harmonization_quantitative}%
    \textbf{Multi-View Lighting Harmonization.}
    We compare our method against ablations on 30 synthetic test scenes.
}
\resizebox{0.8\linewidth}{!}{
\begin{tabular}{l
    S[table-format=2.2, round-mode=places, round-precision=2, detect-weight=true, mode=text]
    S[table-format=1.3, round-mode=places, round-precision=3, detect-weight=true, mode=text]
    S[table-format=1.3, round-mode=places, round-precision=3, detect-weight=true, mode=text]}
\toprule
\bf Method & {\bf PSNR $\uparrow$} & {\bf SSIM $\uparrow$} & {\bf LPIPS $\downarrow$} \\
\midrule
\paper-SV & 25.137302 & 0.807102 & 0.148779 \\
\paper-MV-Edit & 26.371356 & 0.793793 & 0.136203 \\
\paper-MV & \bfseries 30.762927 & \bfseries 0.866890 & \bfseries 0.090679 \\
\bottomrule
\end{tabular}
}
\end{table}

\input{assets/single_image_comparison.tex}

\subsection{Real-time Remixable Lighting}

\Cref{fig:relightable_3dgs} shows our real-time remixable lighting results.
Prior works do not provide the necessary level of fine-grained lighting control for interactive 3D scenes.
Approaches like NeRF-W \cite{MartiRSBDD2021} and Splatfacto-W \cite{XuKK2024} can relight scenes at a per-image level when trained on multi-illumination captures, but lack per-light control.
Text-based editing methods like Instruct-NeRF2NeRF \cite{HaqueTEHK2023} offer general scene editing but are also too imprecise.
In contrast, our method enables interactive control over individual light sources with real-time feedback.
Please see our supplementary material for our video results.

\section{Conclusions and Limitations}
\label{sec:conclusion}

We introduced \paper, the first method for multi-view lighting decomposition and remixing for indoor scenes.
Our approach combines a single-image lighting decomposition model with multi-view harmonization to produce view-consistent per-light decompositions from arbitrary multi-view captures.
By training on a large-scale synthetic dataset with ground-truth OLAT decompositions, our models learn to isolate individual light contributions and enable flexible relighting control.
We further introduce a real-time remixable lighting representation based on 3D Gaussian splatting that allows users to interactively manipulate individual light sources, adjusting their intensities and colors from any viewpoint.
Our method opens new possibilities for interactive scene editing, virtual production, and immersive experiences, where fine-grained control over lighting is essential for achieving photorealistic results.
We believe this work establishes a foundation for future research in controllable multi-view scene lighting.

While introducing a significant step towards consistent, multi-view scene relighting, our approach has remaining limitations.
First, our models are trained exclusively on static synthetic indoor scenes and may not generalize well to outdoor or dynamic scenes.
Second, the limited diversity of light sources in our training data introduces some bias in the lighting decomposition, tending to favor light cones over more diffuse lighting configurations.
Third, distant global illumination editing via HDRIs is not supported by our model, which is left for future work.
Please see the supplement for failure cases and a comparison of single-view versus multi-view consistency.

\iftoggle{cvprfinal}{%
\paragraph{Acknowledgements}
We are grateful to Armen Avetisyan and Samir Aroudj for their support with the Aria Synthetic Environments dataset, and
we are thankful to Jensen Zhou, Samuel Rota Bulò, Yehonathon Litman, and Yuxuan Xue for helpful discussions and feedback.
Nandita Vijaykumar was supported by the Natural Sciences and Engineering Research Council of Canada (NSERC). 
}{}  %

{
    \small
    \bibliographystyle{ieeenat_fullname}
    \bibliography{LuxRemix-CR}
}

\input{LuxRemix-supplement.tex}
\end{document}

%% file: preamble.tex
\usepackage{multirow}
\usepackage{siunitx}
\usepackage{tikz}
\usepackage{quoting,xparse}
\usetikzlibrary{arrows.meta,calc}

\definecolor{figlineblue}{rgb}{0.25, 0.5, 1.0}
\newcommand{\drawrectangleblue}[2]{
	\draw[line width=1.0pt, figlineblue] #1 rectangle #2;
}

\newcommand{\new}[1]{{\color{blue}#1}}
\newcommand{\red}[1]{{\color{red}#1}}

\newcommand{\TODO}[1]{\textbf{\color{red}[TODO: #1]}}

\newcommand{\RL}[1]{\textcolor{green!70!black}{[\textbf{RL:} #1]}}
\newcommand{\CR}[1]{\textcolor{cyan}{[\textbf{CR:} #1]}}
\newcommand{\NM}[1]{\textcolor{blue}{[\textbf{NM:} #1]}}
\newcommand{\NV}[1]{\textcolor{purple!70}{[\textbf{NV:} #1]}}
\newcommand{\ethan}[1]{\textcolor{orange}{ETHAN: #1}}

\definecolor{placeholder}{rgb}{0.6,0.8,0.95}

\renewcommand{\TODO}[1]{}

\renewcommand{\new}[1]{}
\renewcommand{\red}[1]{}
\renewcommand{\RL}[1]{}
\renewcommand{\CR}[1]{}
\renewcommand{\NM}[1]{}
\renewcommand{\NV}[1]{}
\renewcommand{\ethan}[1]{}

\usepackage{silence}
\usepackage{microtype}
\makeatletter
\renewcommand\paragraph{\@startsection{paragraph}{4}{\z@}%
	{0.75ex \@plus.5ex \@minus.2ex}%
	{-1em}%
	{\normalfont\normalsize\bfseries\maybe@addperiod}}
\newcommand{\maybe@addperiod}[1]{#1\@addpunct{.}}
\makeatother

\usepackage{xspace}
\newcommand{\paper}{LuxRemix\xspace}

%% file: assets/synthetic_data_example.tex
\begin{figure}[t]
  \centering
  \begin{tikzpicture}[
    label/.style={anchor=north west, fill=black, fill opacity=0.6, text=white, text opacity=1, inner sep=2pt, font=\sffamily\footnotesize}
  ]
    \def\imgwidth{0.49\linewidth}
    \def\colsep{2pt}
    \def\rowsep{2pt}

    \def\scene{014658}
    
    \node[inner sep=0pt, anchor=north west] (img11) at (0,0) {%
      \includegraphics[width=\imgwidth]{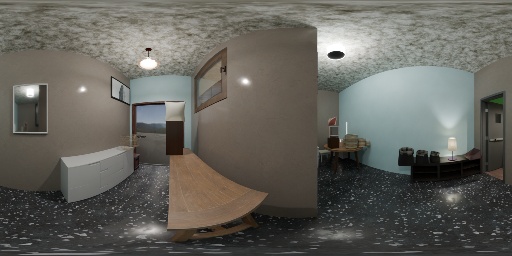}%
    };
    \node[inner sep=0pt, anchor=north west] (img12) at ([xshift=\colsep]img11.north east) {%
      \includegraphics[width=\imgwidth]{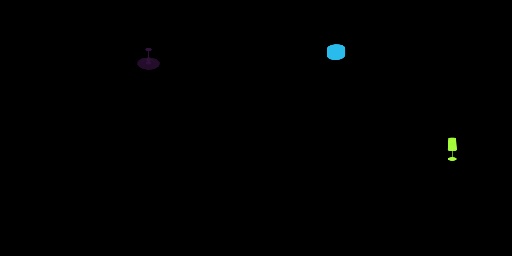}%
    };
    
    \node[inner sep=0pt, anchor=north west] (img21) at ([yshift=-\rowsep]img11.south west) {%
      \includegraphics[width=\imgwidth]{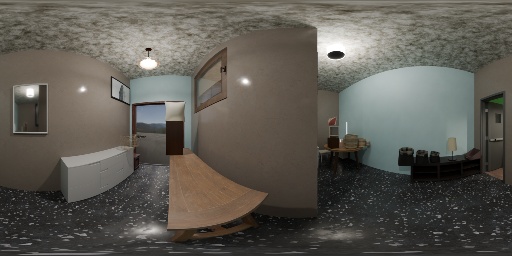}%
    };
    \node[inner sep=0pt, anchor=north west] (img22) at ([xshift=\colsep]img21.north east) {%
      \includegraphics[width=\imgwidth]{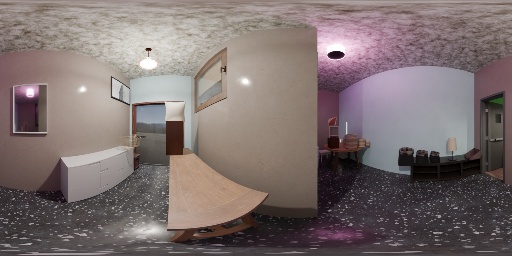}%
    };
    
    \node[inner sep=0pt, anchor=north west] (img31) at ([yshift=-\rowsep]img21.south west) {%
      \includegraphics[width=\imgwidth]{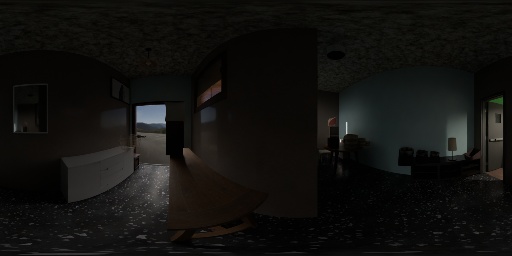}%
    };
    \node[inner sep=0pt, anchor=north west] (img32) at ([xshift=\colsep]img31.north east) {%
      \includegraphics[width=\imgwidth]{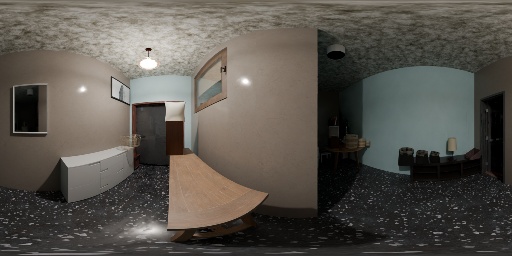}%
    };
    
    \node[inner sep=0pt, anchor=north west] (img41) at ([yshift=-\rowsep]img31.south west) {%
      \includegraphics[width=\imgwidth]{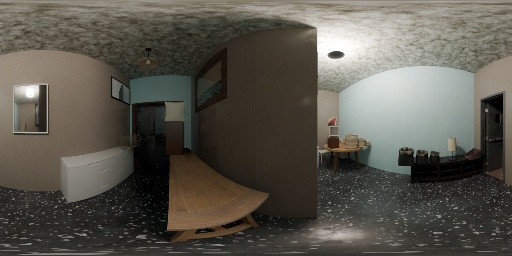}%
    };
    \node[inner sep=0pt, anchor=north west] (img42) at ([xshift=\colsep]img41.north east) {%
      \includegraphics[width=\imgwidth]{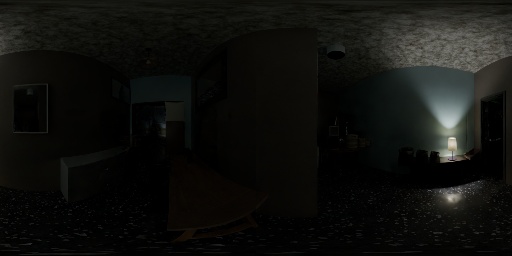}%
    };
    
    \node[label] at (img11.north west) {Fully lit scene};
    \node[label] at (img12.north west) {Light mask};
    \node[label] at (img21.north west) {Light config 1};
    \node[label] at (img22.north west) {Light config 2};
    \node[label] at (img31.north west) {Ambient lighting};
    \node[label] at (img32.north west) {OLAT 1};
    \node[label] at (img41.north west) {OLAT 2};
    \node[label] at (img42.north west) {OLAT 3};
  \end{tikzpicture}
  \caption{\label{fig:synthetic_data_example}%
    \textbf{Synthetic multi-light data example.}
    Our synthetic dataset comprises over 12,000 indoor scenes with procedurally generated light sources, each rendered as equirectangular images under multiple lighting conditions: fully lit, randomly lit, ambient only, and one-light-at-a-time (OLAT).
    During training, we sample perspective views from these equirectangular images on the fly.
  }
\vspace{-0.5em}
\end{figure}

%% file: assets/single_image_light_editing.tex
\begin{figure*}[t]
  \centering
  \begin{tikzpicture}[
    label/.style={anchor=north west, fill=black, fill opacity=0.6, text=white, text opacity=1, inner sep=2pt, font=\sffamily\footnotesize}
  ]
    \def\imgwidth{80pt}
    \def\imgsep{2pt}
    
    \node[inner sep=0pt, anchor=north west] (input) at (0,0) {%
      \includegraphics[width=161.5pt]{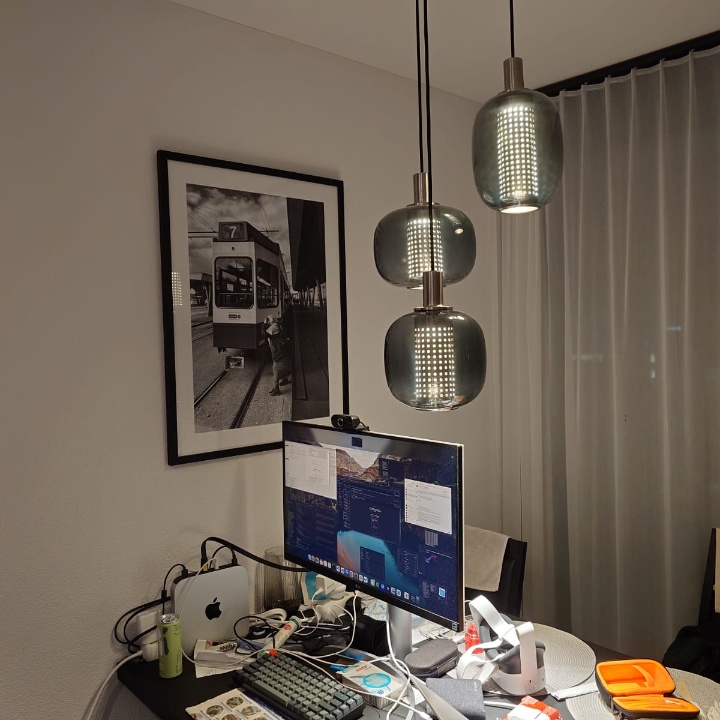}%
    };
    
    \node[inner sep=0pt, anchor=north west] (mask1) at ([xshift=\imgsep]input.north east) {%
      \includegraphics[width=\imgwidth]{assets/single_image_light_editing/table_1/mask_2}%
    };
    \node[inner sep=0pt, anchor=north west] (mask2) at ([xshift=\imgsep]mask1.north east) {%
      \includegraphics[width=\imgwidth]{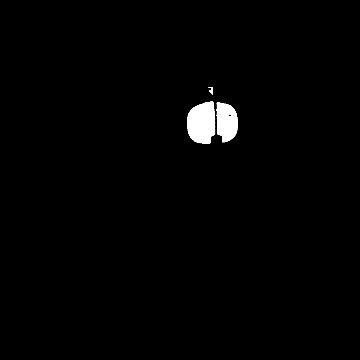}%
    };
    \node[inner sep=0pt, anchor=north west] (mask3) at ([xshift=\imgsep]mask2.north east) {%
      \includegraphics[width=\imgwidth]{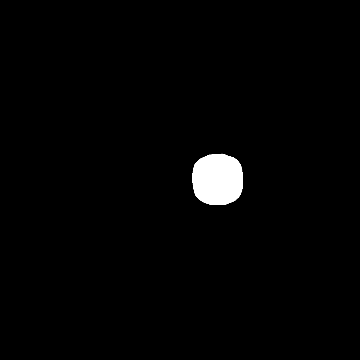}%
    };
    \node[inner sep=0pt, anchor=north west] (mask4) at ([xshift=\imgsep]mask3.north east) {%
      \includegraphics[width=\imgwidth]{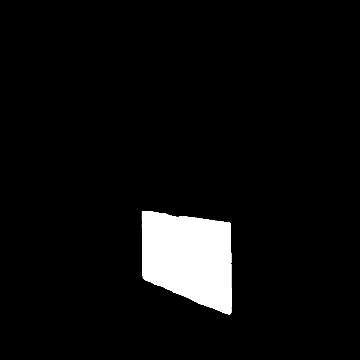}%
    };
    
    \node[inner sep=0pt, anchor=north west] (olat1) at ([yshift=-1pt]mask1.south west) {%
      \includegraphics[width=\imgwidth]{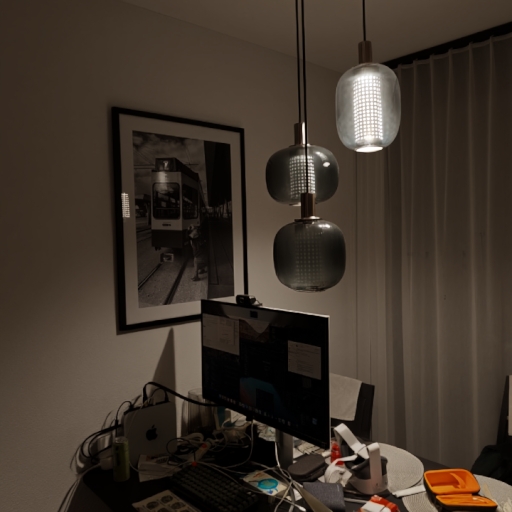}%
    };
    \node[inner sep=0pt, anchor=north west] (olat2) at ([yshift=-1pt]mask2.south west) {%
      \includegraphics[width=\imgwidth]{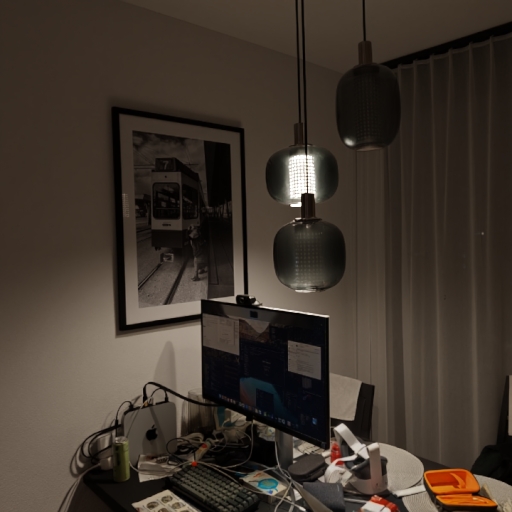}%
    };
    \node[inner sep=0pt, anchor=north west] (olat3) at ([yshift=-1pt]mask3.south west) {%
      \includegraphics[width=\imgwidth]{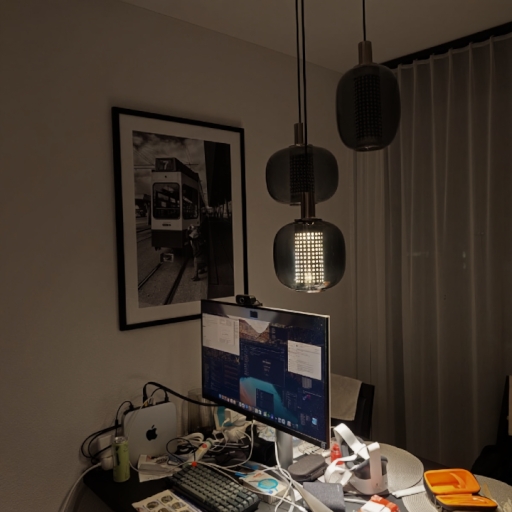}%
    };
    \node[inner sep=0pt, anchor=north west] (olat4) at ([yshift=-1pt]mask4.south west) {%
      \includegraphics[width=\imgwidth]{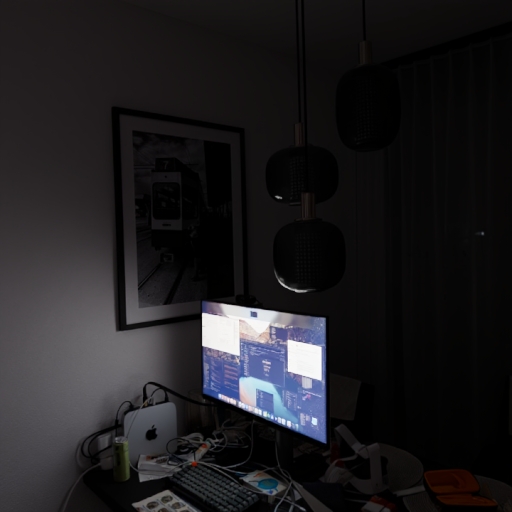}%
    };

    \node[label] at (input.north west) {Input};
    \node[label] at (mask1.north west) {Mask 1};
    \node[label] at (mask2.north west) {Mask 2};
    \node[label] at (mask3.north west) {Mask 3};
    \node[label] at (mask4.north west) {Mask 4};
    \node[label] at (olat1.north west) {OLAT 1};
    \node[label] at (olat2.north west) {OLAT 2};
    \node[label] at (olat3.north west) {OLAT 3};
    \node[label] at (olat4.north west) {OLAT 4};
  \end{tikzpicture}
  \caption{\label{fig:single_image_light_editing}%
    \textbf{Single-image Lighting Decomposition.}
    Given the light masks in the top row, our single-image lighting decomposition model realistically decomposes the scene lighting in the input image (on the left) into four individual one-light-at-a-time (OLAT) light sources.
  }
\end{figure*}

%% file: assets/lighting_harmonization.tex
\begin{figure*}[t]
  \centering
  \begin{tikzpicture}
    \node[anchor=south west, inner sep=0] (image) at (0,0) {
        \includegraphics[width=0.97\linewidth]{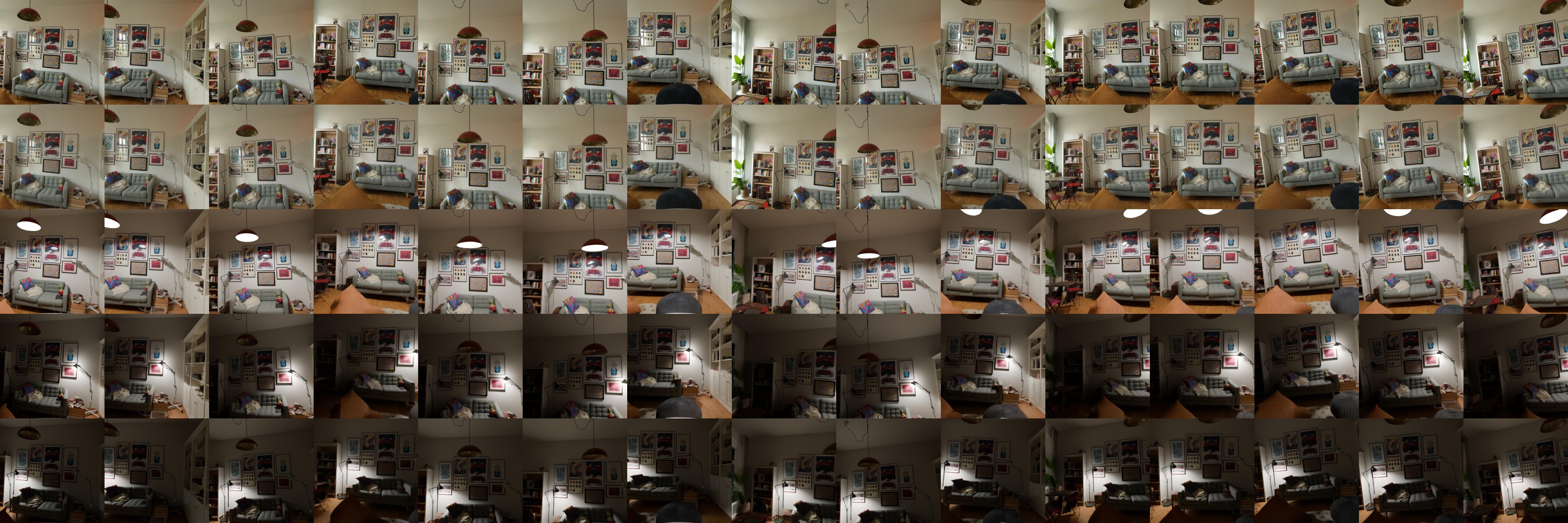}
    };
    \begin{scope}[
        x={($0.1*(image.south east)$)},
        y={($0.1*(image.north west)$)},
        font=\sffamily\footnotesize,
        nodes={text depth=0.25ex,text height=1.25ex}
        ]
        \drawrectangleblue{(0, 0)}{(0.67, 8)}
        \node[above,rotate=90] at (0.02, 9) {Inputs};
        \node[above,rotate=90] at (0.02, 7) {Ambient};
        \node[above,rotate=90] at (0.02, 5) {OLAT 1};
        \node[above,rotate=90] at (0.02, 3) {OLAT 2};
        \node[above,rotate=90] at (0.02, 1) {OLAT 3};
    \end{scope}
  \end{tikzpicture}\\[2pt]
  \begin{tikzpicture}
    \node[anchor=south west, inner sep=0] (image) at (0,0) {
        \includegraphics[width=0.97\linewidth]{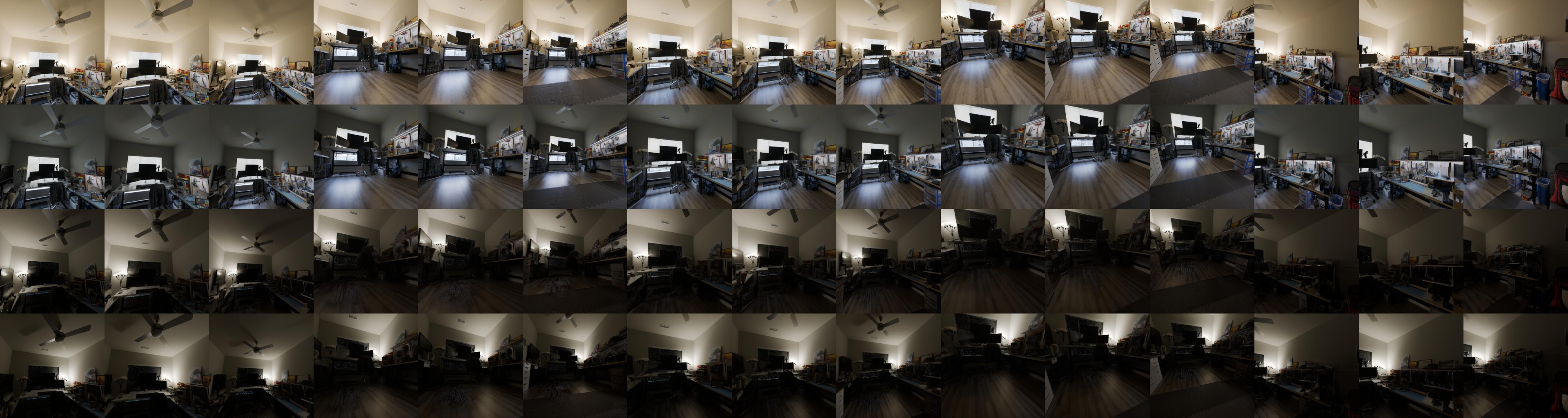}
    };
    \begin{scope}[
        x={($0.1*(image.south east)$)},
        y={($0.1*(image.north west)$)},
        font=\sffamily\footnotesize,
        nodes={text depth=0.25ex,text height=1.25ex}
        ]
        \drawrectangleblue{(0, 0)}{(0.67, 7.5)}
        \node[above,rotate=90] at (0.02, 8.75) {Inputs};
        \node[above,rotate=90] at (0.02, 6.25) {Ambient};
        \node[above,rotate=90] at (0.02, 3.75) {OLAT 1};
        \node[above,rotate=90] at (0.02, 1.25) {OLAT 2};
    \end{scope}
  \end{tikzpicture}

  \caption{\label{fig:lighting_harmonization}%
    \textbf{Multi-view Lighting Harmonization.}
    From the first input image, we decompose scene lighting into ambient lighting and per-light OLAT components (highlighted in blue), then propagate these consistently across all views (top row).
    For the top scene, \paper turns on all three lamps individually; for the bottom scene, it turns off either lamp.
    Real-world captures from Zip-NeRF \cite{BarroMVSH2023} and VR-NeRF \cite{XuALGBKRPKBLZR2023}.
  }
\end{figure*}

%% file: assets/single_image_comparison.tex
\begin{figure*}[t]
  \centering
  \begin{tikzpicture}[font=\sffamily\footnotesize]
    \def\imgwidth{80pt}
    \def\imgsep{2pt}
    
    \node[inner sep=0pt, anchor=north west] (img1-input) at (0,0) {%
      \includegraphics[width=\imgwidth]{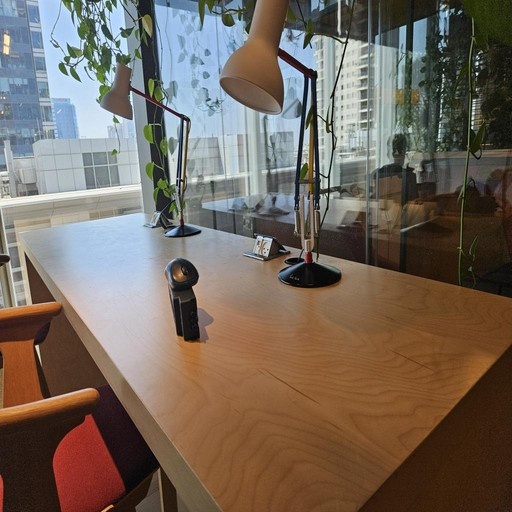}%
    };
    \node[inner sep=0pt, anchor=north west] (img1-scribblelight) at ([xshift=\imgsep]img1-input.north east) {%
      \includegraphics[width=\imgwidth]{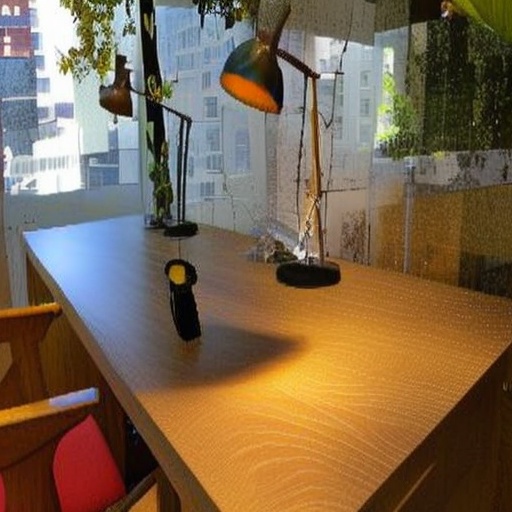}%
    };
    \node[inner sep=0pt, anchor=north west] (img1-qwenimage) at ([xshift=\imgsep]img1-scribblelight.north east) {%
      \includegraphics[width=\imgwidth]{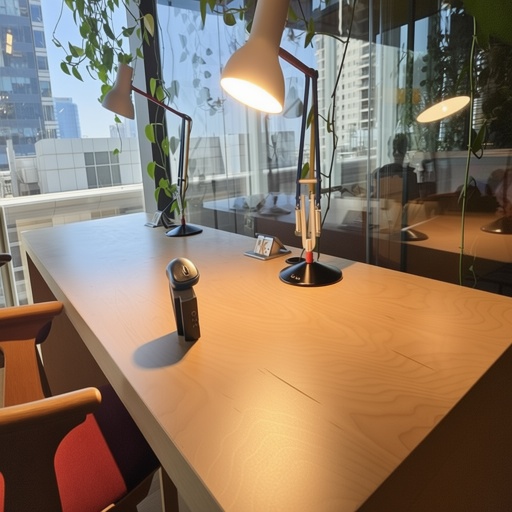}%
    };
    \draw[red, line width=.8mm] ([xshift=69pt, yshift=-17pt]img1-qwenimage.north west) circle (8pt);
    \node[inner sep=0pt, anchor=north west] (img1-lightlab) at ([xshift=\imgsep]img1-qwenimage.north east) {%
      \includegraphics[width=\imgwidth]{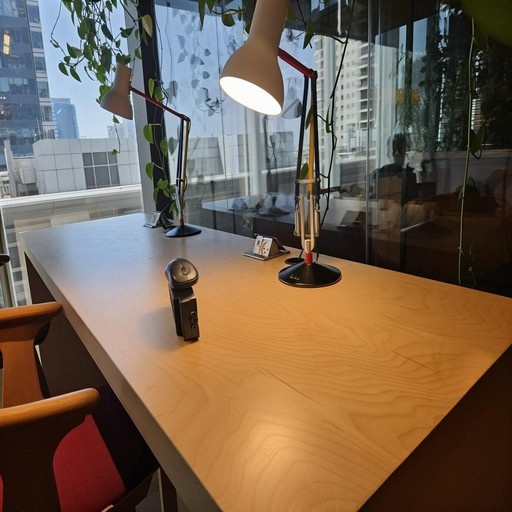}%
    };
    \node[inner sep=0pt, anchor=north west] (img1-ours) at ([xshift=\imgsep]img1-lightlab.north east) {%
      \includegraphics[width=\imgwidth]{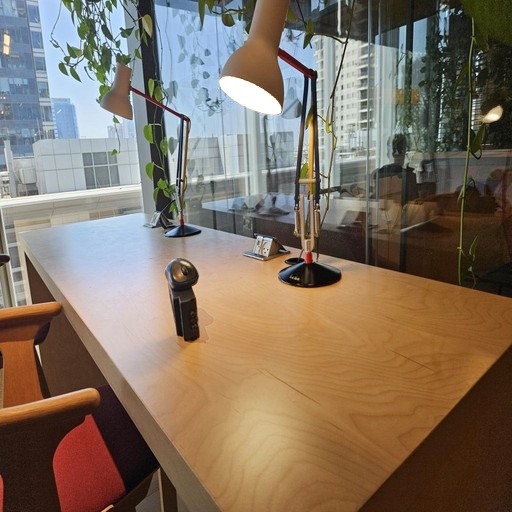}%
    };
    \node[inner sep=0pt, anchor=north west] (img1-gt) at ([xshift=\imgsep]img1-ours.north east) {%
      \includegraphics[width=\imgwidth]{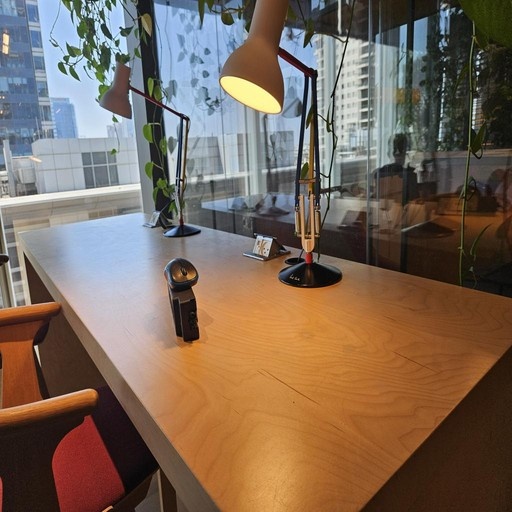}%
    };

    \node[inner sep=0pt, anchor=north, below=\imgsep] (img2-input) at (img1-input.south) {%
      \includegraphics[width=\imgwidth]{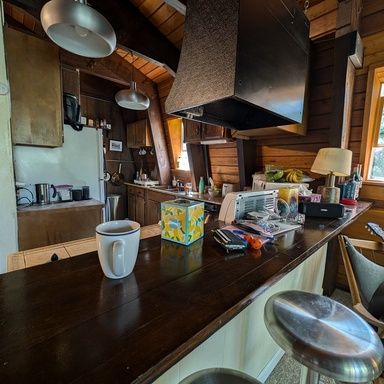}%
    };
    \node[inner sep=0pt, anchor=north west] (img2-scribblelight) at ([xshift=\imgsep]img2-input.north east) {%
      \includegraphics[width=\imgwidth]{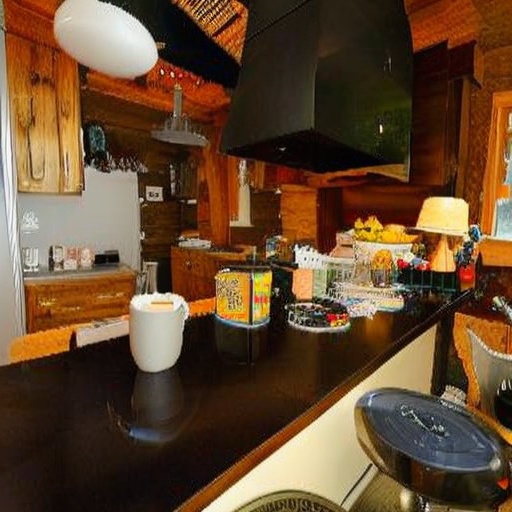}%
    };
    \node[inner sep=0pt, anchor=north west] (img2-qwenimage) at ([xshift=\imgsep]img2-scribblelight.north east) {%
      \includegraphics[width=\imgwidth]{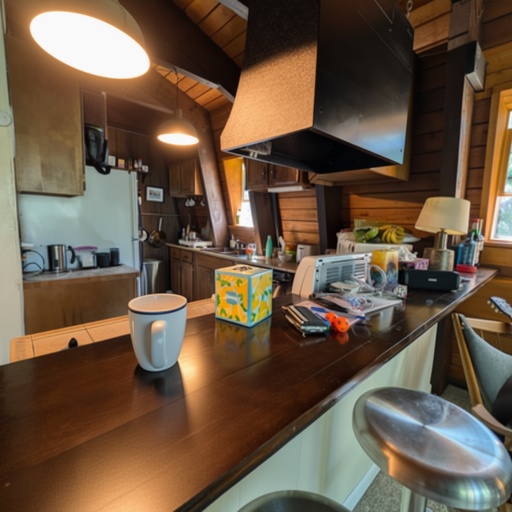}%
    };
    \node[inner sep=0pt, anchor=north west] (img2-lightlab) at ([xshift=\imgsep]img2-qwenimage.north east) {%
      \includegraphics[width=\imgwidth]{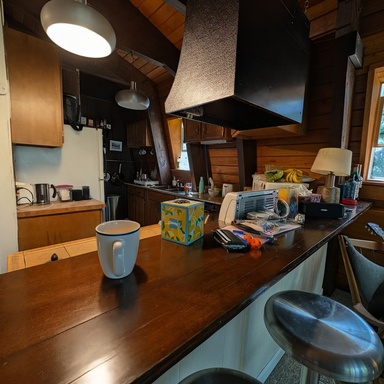}%
    };
    \draw[-stealth, red, line width=1.5mm] ([xshift=50pt, yshift=-70pt]img2-lightlab.north west) -- ([xshift=30pt, yshift=-70pt]img2-lightlab.north west);
    \node[inner sep=0pt, anchor=north west] (img2-ours) at ([xshift=\imgsep]img2-lightlab.north east) {%
      \includegraphics[width=\imgwidth]{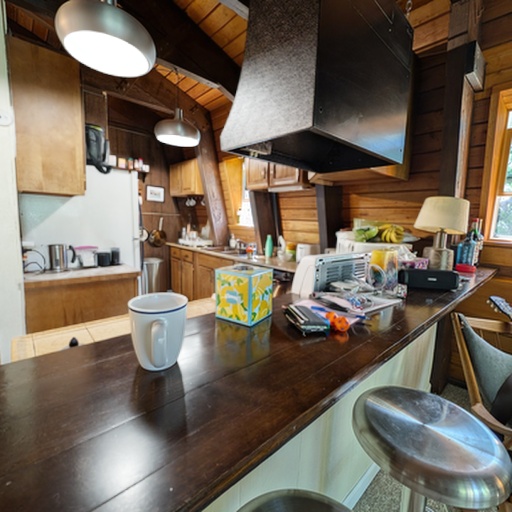}%
    };
    \node[inner sep=0pt, anchor=north west] (img2-gt) at ([xshift=\imgsep]img2-ours.north east) {%
      \includegraphics[width=\imgwidth]{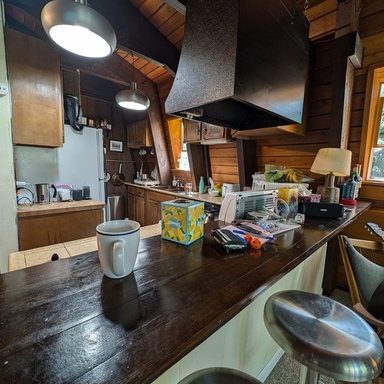}%
    };

    \node[inner sep=0pt, anchor=north, below=\imgsep] (img4-input) at (img2-input.south) {%
      \includegraphics[width=\imgwidth]{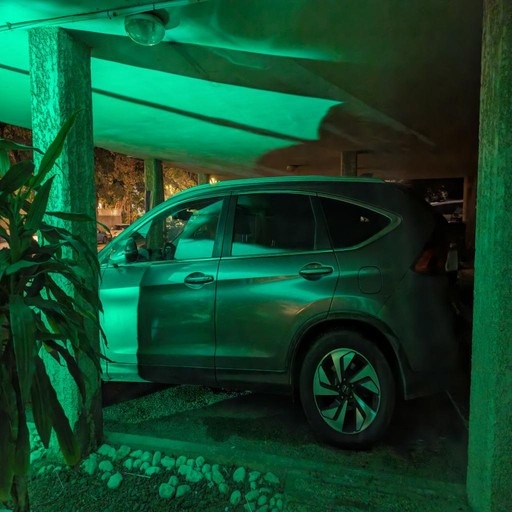}%
    };
    \node[inner sep=0pt, anchor=north west] (img4-scribblelight) at ([xshift=\imgsep]img4-input.north east) {%
      \includegraphics[width=\imgwidth]{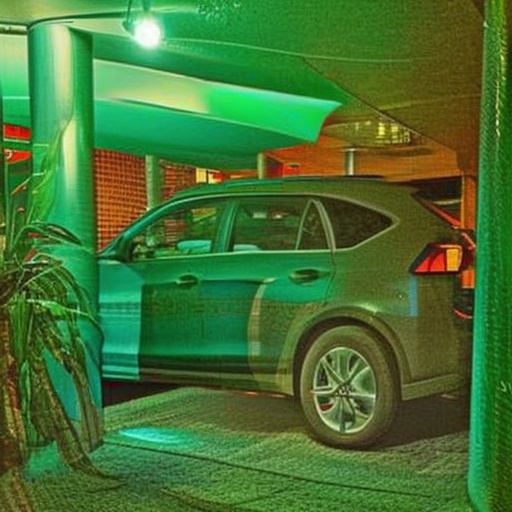}%
    };
    \node[inner sep=0pt, anchor=north west] (img4-qwenimage) at ([xshift=\imgsep]img4-scribblelight.north east) {%
      \includegraphics[width=\imgwidth]{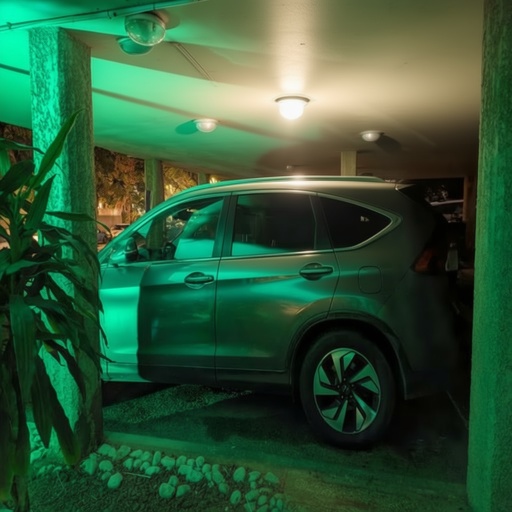}%
    };
    \draw[red, line width=.8mm] ([xshift=24pt, yshift=-5pt]img4-qwenimage.north west) circle (6pt);
    \draw[red, line width=.8mm] ([xshift=32pt, yshift=-19pt]img4-qwenimage.north west) circle (6pt);
    \draw[red, line width=.8mm] ([xshift=45pt, yshift=-17pt]img4-qwenimage.north west) circle (6pt);
    \draw[red, line width=.8mm] ([xshift=58pt, yshift=-21pt]img4-qwenimage.north west) circle (6pt);

    \node[inner sep=0pt, anchor=north west] (img4-lightlab) at ([xshift=\imgsep]img4-qwenimage.north east) {%
      \includegraphics[width=\imgwidth]{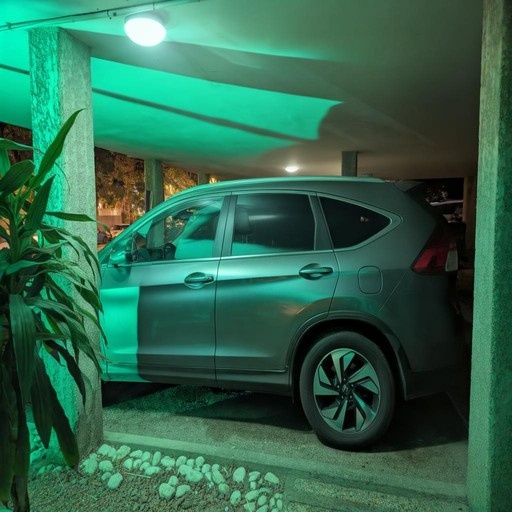}%
    };
    \node[inner sep=0pt, anchor=north west] (img4-ours) at ([xshift=\imgsep]img4-lightlab.north east) {%
      \includegraphics[width=\imgwidth]{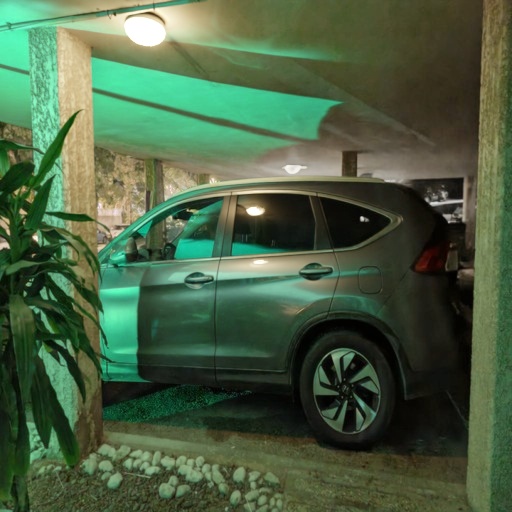}%
    };
    \node[inner sep=0pt, anchor=north west] (img4-gt) at ([xshift=\imgsep]img4-ours.north east) {%
      \includegraphics[width=\imgwidth]{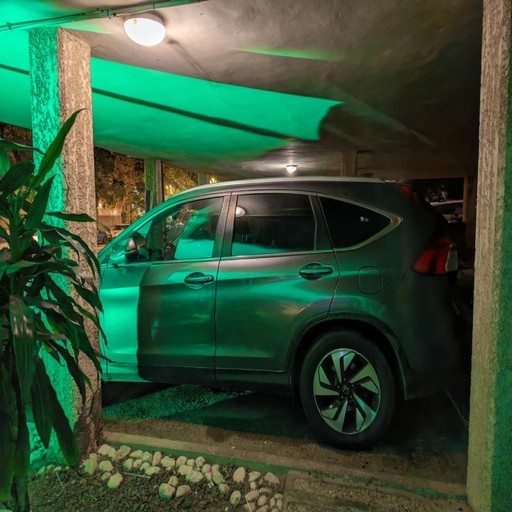}%
    };
    
    \node[inner sep=0pt, anchor=north, below=\imgsep] (img5-input) at (img4-input.south) {%
      \includegraphics[width=\imgwidth]{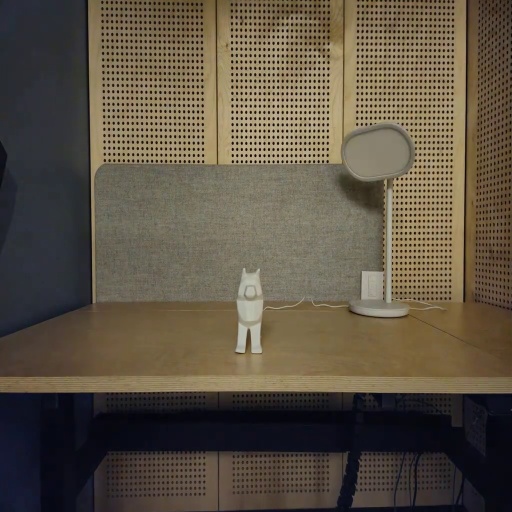}%
    };
    \node[inner sep=0pt, anchor=north west] (img5-scribblelight) at ([xshift=\imgsep]img5-input.north east) {%
      \includegraphics[width=\imgwidth]{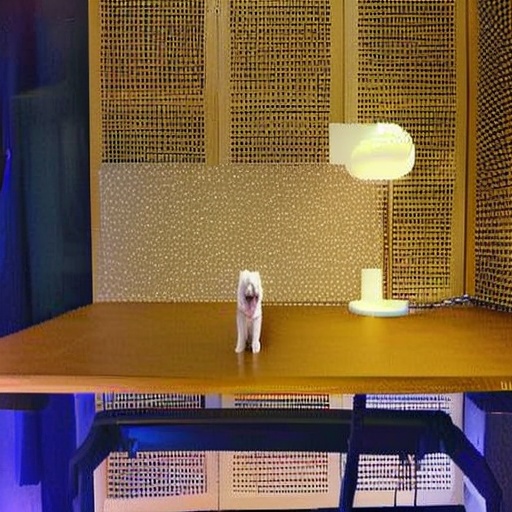}%
    };
    \node[inner sep=0pt, anchor=north west] (img5-qwenimage) at ([xshift=\imgsep]img5-scribblelight.north east) {%
      \includegraphics[width=\imgwidth]{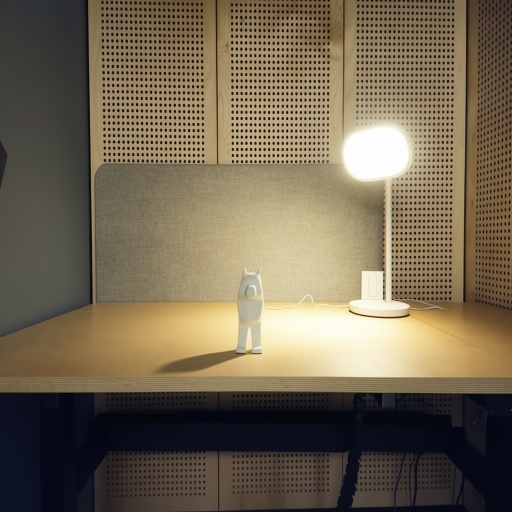}%
    };
    \node[inner sep=0pt, anchor=north west] (img5-lightlab) at ([xshift=\imgsep]img5-qwenimage.north east) {%
      \includegraphics[width=\imgwidth]{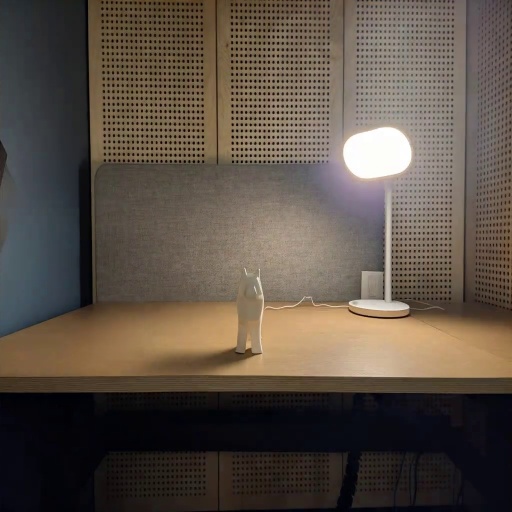}%
    };
    \node[inner sep=0pt, anchor=north west] (img5-ours) at ([xshift=\imgsep]img5-lightlab.north east) {%
      \includegraphics[width=\imgwidth]{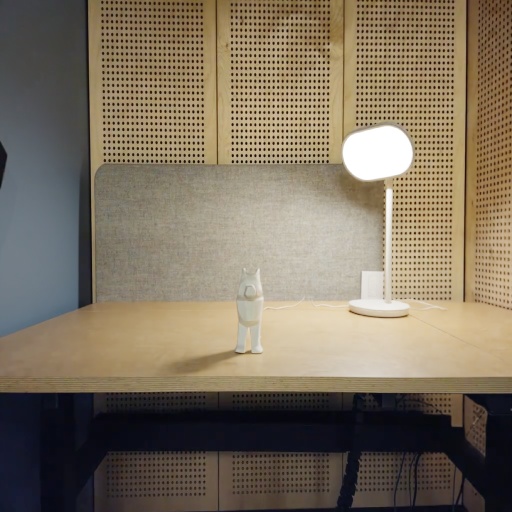}%
    };
    \node[inner sep=0pt, anchor=north west, fill=gray!40, minimum width=\imgwidth, minimum height=\imgwidth] (img5-gt) at ([xshift=\imgsep]img5-ours.north east) {\Large N/A};
    
    \node[inner sep=0pt, anchor=north, below=4pt] (img6-input) at (img5-input.south) {%
      \includegraphics[width=\imgwidth]{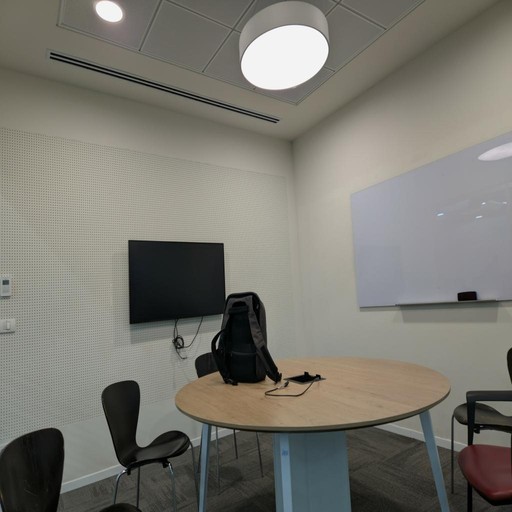}%
    };
    \node[inner sep=0pt, anchor=north west] (img6-scribblelight) at ([xshift=\imgsep]img6-input.north east) {%
      \includegraphics[width=\imgwidth]{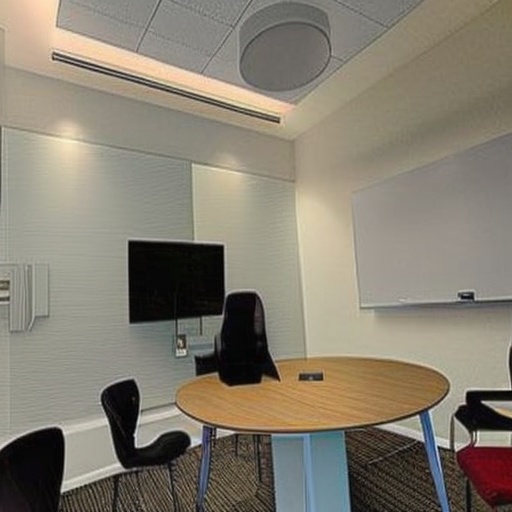}%
    };
    \node[inner sep=0pt, anchor=north west] (img6-qwenimage) at ([xshift=\imgsep]img6-scribblelight.north east) {%
      \includegraphics[width=\imgwidth]{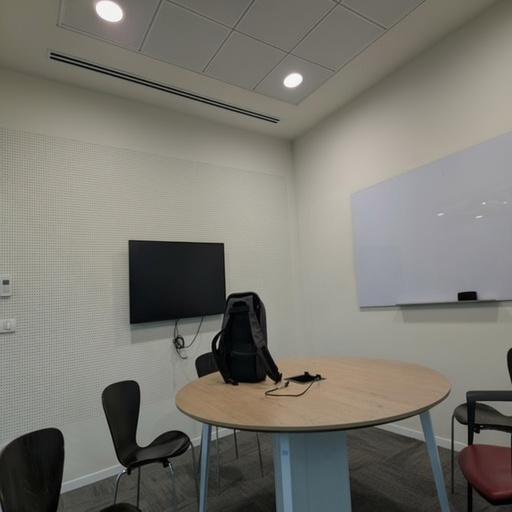}%
    };
    \draw[-stealth, red, line width=1.5mm] ([xshift=70pt, yshift=-10pt]img6-qwenimage.north west) -- ([xshift=50pt, yshift=-7pt]img6-qwenimage.north west);
    \node[inner sep=0pt, anchor=north west] (img6-lightlab) at ([xshift=\imgsep]img6-qwenimage.north east) {%
      \includegraphics[width=\imgwidth]{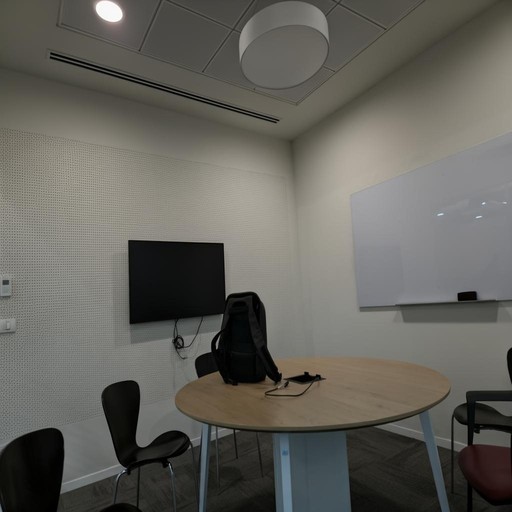}%
    };
    \node[inner sep=0pt, anchor=north west] (img6-ours) at ([xshift=\imgsep]img6-lightlab.north east) {%
      \includegraphics[width=\imgwidth]{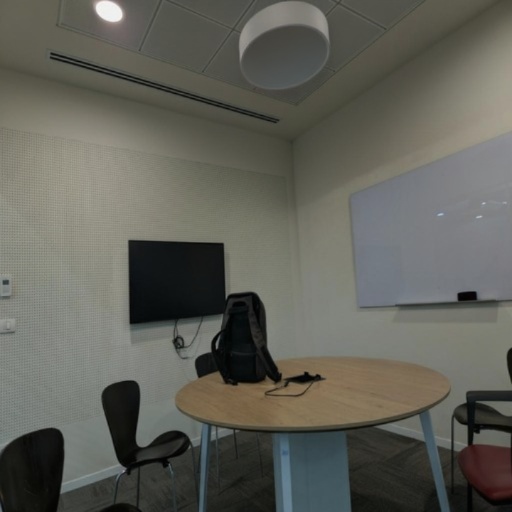}%
    };
    \node[inner sep=0pt, anchor=north west] (img6-gt) at ([xshift=\imgsep]img6-ours.north east) {%
      \includegraphics[width=\imgwidth]{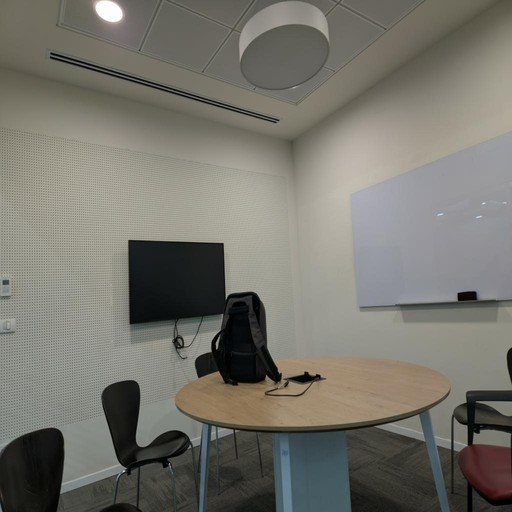}%
    };

    \node[anchor=south, text depth=0.25ex, inner sep=2pt] at ([yshift=2pt]img1-input.north) {Input Image};
    \node[anchor=south, text depth=0.25ex, inner sep=2pt] at ([yshift=2pt]img1-scribblelight.north) {ScribbleLight \cite{ChoiWPBS2025}};
    \node[anchor=south, text depth=0.25ex, inner sep=2pt] at ([yshift=2pt]img1-qwenimage.north) {Qwen-Image \cite{WuLZLGYYBXCCTZWYYCLLZMWNCCPQWWYWFXWZZWCL2025}};
    \node[anchor=south, text depth=0.25ex, inner sep=2pt] at ([yshift=2pt]img1-lightlab.north) {LightLab \cite{MagarHTPRSH2025}};
    \node[anchor=south, text depth=0.25ex, inner sep=2pt] at ([yshift=2pt]img1-ours.north) {\paper-SV (ours)};
    \node[anchor=south, text depth=0.25ex, inner sep=2pt] at ([yshift=2pt]img1-gt.north) {Ground Truth};
    
  \end{tikzpicture}
  \caption{\label{fig:single_image_comparison}%
    \textbf{Comparison of Single-Image Light Editing.}
    Given an input image and light mask, we compare single-image light editing results across methods.
    ScribbleLight \cite{ChoiWPBS2025} fails to produce plausible lighting.
    Qwen-Image \cite{WuLZLGYYBXCCTZWYYCLLZMWNCCPQWWYWFXWZZWCL2025} shows 
    great zero-shot image editing capabilities, but sometimes spuriously adds or 
    removes lights and reflections without mask conditioning.
    LightLab \cite{MagarHTPRSH2025} occasionally struggles with reflections.
    Our \paper-SV model produces convincing results throughout.
    Real-world images from LightLab \cite{MagarHTPRSH2025}.
  }
\end{figure*}

%% file: LuxRemix-supplement.tex
\clearpage
\maketitlesupplementary

\noindent
This supplemental document provides additional details for \paper.
We first describe the data preparation process in \cref{sec:data}, then introduce additional details about our models in \cref{sec:models}, and finally provide additional results and comparisons in \cref{sec:additional_results}.

Please see our project website \href{https://luxremix.github.io/}{https://luxremix.github.io} for the following additional supplemental material:
\begin{itemize}
    \item \textbf{Interactive Lighting Blend Explorer:}\\
    A browser-based demo for interactive single-image light editing based on our lighting decompositions.

    \item \textbf{Single-Image Lighting Decomposition:}\\
    Shows the decomposition of seven light sources and their flexible recombination for various relighting effects.

    \item \textbf{Multi-View Lighting Harmonization:}\\
    Visualizes the decomposition and harmonization of multiple light sources across multiple views.

    \item \textbf{Real-time Remixable Lighting:}\\
    Screen recordings from our modified Splatfacto renderer in the Nerfstudio viewer demonstrating our interactive relighting with 3D Gaussian splatting.
    Note that temporarily lower rendering resolutions are due to the dynamic resolution rendering implemented in the Nerfstudio viewer.
\end{itemize}

\section{Data}
\label{sec:data}

Here we provide additional information for how we created the training data for \paper.
We render scenes from \citet{AvetiXHYAPZFHOEMNB2024} with procedurally generated light sources using Infinigen \cite{RaistMKYZHWPALMD2024} into many one-light-at-a-time (OLAT) equirectangular images.
We choose equirectangular 360° images to save on the rendering cost and allow dynamic sampling of diverse perspective camera viewpoints on the fly while training the model.
Please see \cref{tab:render_storage} for a comparison of the rendering and storage costs.

\begin{table}[h!]
    \centering
    \caption{\label{tab:render_storage}%
        \textbf{Rendering and storage comparison.}
        The perspective views are rendered at resolution 512$\times$512 pixels, and the equirectangular 360° images are rendered at resolution 2048$\times$1024 pixels.
        64 samples per pixel are used for Blender's Cycles path tracer \cite{BOC2025}.
    }
    \begin{tabular}{lrr}
        \toprule
        \textbf{Setup} & \textbf{Render time} & \textbf{Storage} \\
        \midrule
        20,000 perspective images & 140,000 sec & 17.2\,GB \\
        1,000 360° images         &  27,000 sec &  6.8\,GB \\
                                  & = 19\%      & = 40\% \\
        \bottomrule
    \end{tabular}
\end{table}

\begin{figure*}[t]
    \centering
    \begin{subfigure}{0.4312\linewidth}
        \centering
        \includegraphics[width=\linewidth]{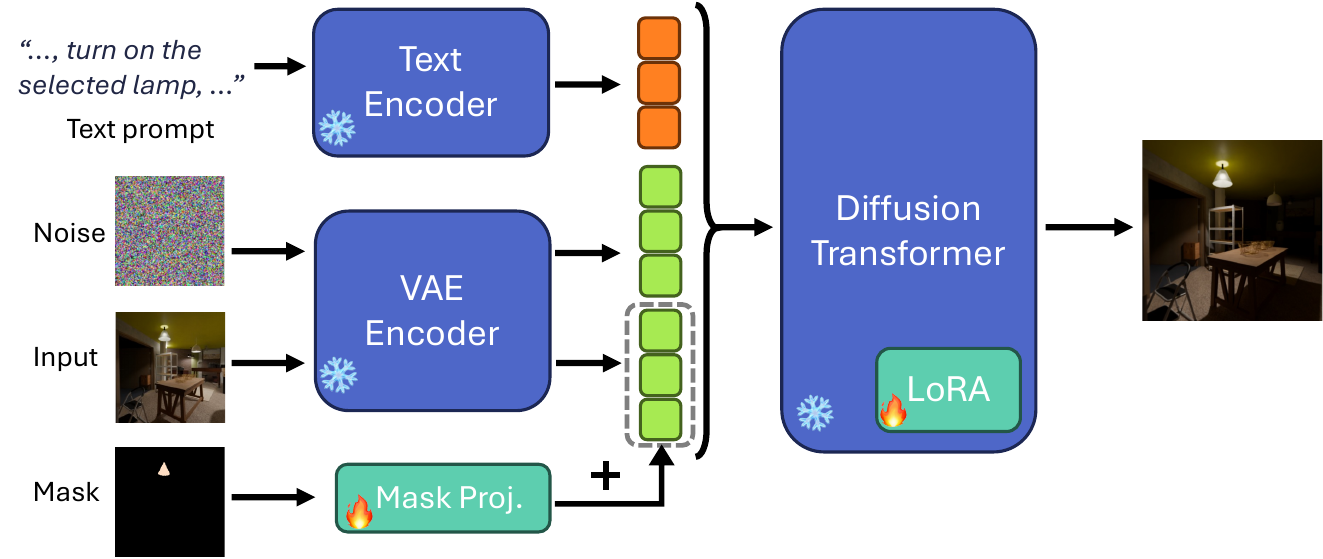}
        \caption{\paper-SV (single-view)}
        \label{fig:sv_model}
    \end{subfigure}
    \hfill
    \begin{subfigure}{0.539\linewidth}
        \centering
        \includegraphics[width=\linewidth]{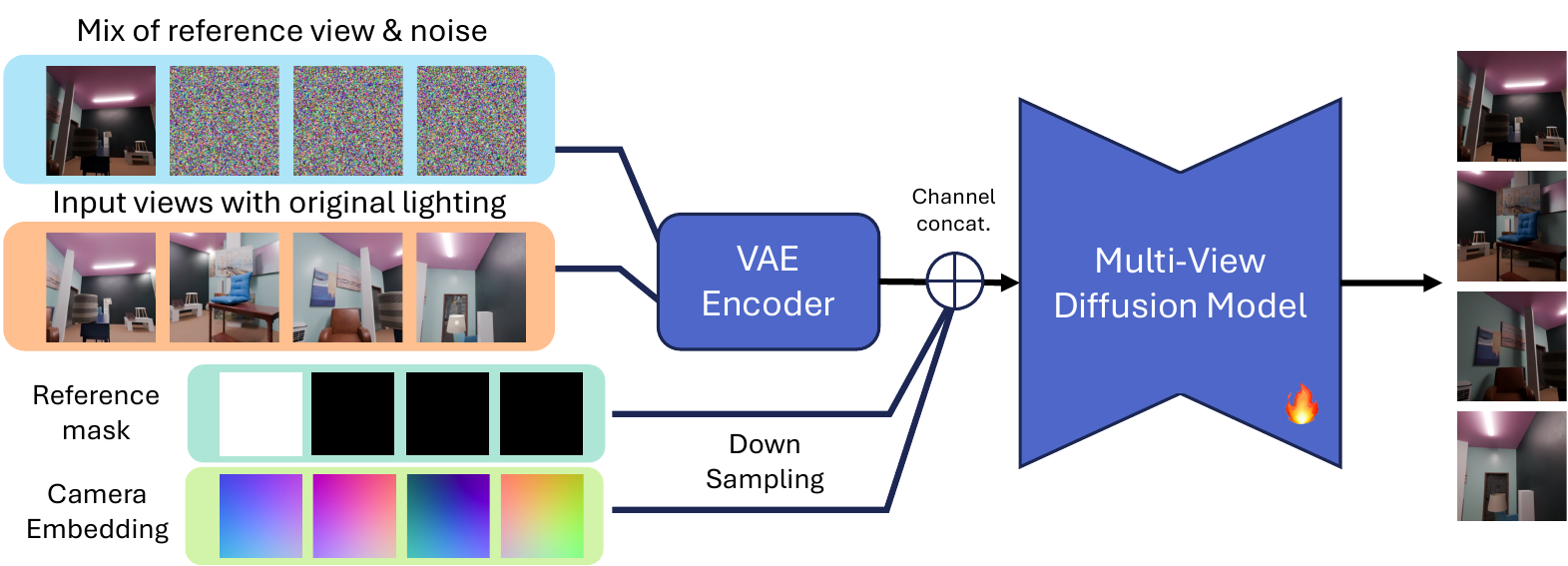}
        \caption{\paper-MV (multi-view)}
        \label{fig:mv_model}
    \end{subfigure}
    \caption{\label{fig:sv_mv_models}%
        \textbf{Model Architectures.} \textbf{(a)} Single-view lighting decomposition and \textbf{(b)} multi-view lighting harmonization.
    }
\end{figure*}

\paragraph{Rendering scenes once}

To reduce our dataset size while maintaining flexibility during training, we store equirectangular 360° images instead of pre-rendering a large collection of perspective views that our model ingests.
Equirectangular representations compactly encode the full scene from a single viewpoint, allowing arbitrary camera perspectives to be sampled on the fly.
In contrast, storing every possible perspective image incurs significant storage overhead and limits viewpoint diversity at training time.
For example, training on 20K pre-rendered perspective images with resolution 512$\times$512 pixels would require 17.2\,GB of storage.
Instead we could store 1K equirectangular images with a resolution of 2048$\times$1024 pixels in just 6.8\,GB and sample these from new viewpoints on the fly, assuming we can always sample at least 20 distinct views per equirectangular image.
This design substantially reduces disk usage and data preparation cost while preserving the ability to generate diverse, view-consistent training inputs dynamically.
We found it particularly useful to render the dataset once and use the stored data for the entire project.

\paragraph{Implementation details}

As described in \cref{sec:data}, we use Blender's Cycles renderer \cite{BOC2025} to generate equirectangular HDR images for each synthetic room.
For each room, we sample four distinct locations to capture equirectangular images at a resolution of 2048$\times$1024 pixels.
Rendering is performed with 64 samples per pixel (spp), using the OptiX denoiser to enhance path-traced results.
We used approximately 2,800 GPU hours on NVIDIA A100s to render the complete dataset.
In total, we rendered approximately 49,600 different equirectangular views across over 12,000 synthetic rooms.
The additional depth map and light mask are rendered for each 360° image to assist effective perspective view sampling.
Combining all different light passes, the total dataset size is about 9\,TB.

\paragraph{On-the-fly sampling}

As our model is designed for regular perspective views, during training, the perspective views are sampled from the rendered equirectangular 360° images.
For single-image light editing, we sample perspective views such that the target light source is visible within the field of view (FOV).
Specifically, we use the equirectangular light mask to select a visible light source, and then use its center to guide the sampling range of azimuth and elevation for the perspective projection.

For multi-view lighting, after selecting an initial perspective view that contains the target light source within the field of view, we sample additional perspective views to incrementally increase the overall coverage of visible regions.
These views are drawn from multiple source equirectangular 360° images and are chosen to ensure that consecutive views share overlapping regions, promoting consistency across frames.
To guide this sampling, we project depth maps from the various equirectangular images into a global coordinate system, allowing us to accurately identify co-visible areas within the room.
Additionally, depth information is used to avoid sampling views that are too close to scene geometry or the camera, ensuring a diverse and meaningful set of training perspectives.

After sampling the perspective views, we generate training data pairs on-the-fly.
For the OLAT decomposition task, we select a target OLAT image $I_\text{target}$ and its corresponding full-light image $I_\text{full}$, forming the training pair $(I_\text{full}, I_\text{target})$.
For the one-light-off task, we compute $I_\text{one-off} = I_\text{full} - \boldsymbol{c}_\text{target} \cdot I_\text{target}$, where $\boldsymbol{c}_\text{target}$ is the scaling factor for the target OLAT image as indicated in the rendering metadata.
The resulting pair $(I_\text{full}, I_\text{one-off})$ teaches the model to turn off the selected light source.
To further augment the training set, we perform additional lighting composition by loading multiple OLAT images and adding them to the full-light image.
This produces new, well-lit image variants such as $I_\text{full}^\prime = I_\text{full} + \sum_{i=1}^{N} \boldsymbol{c}_i^\prime \cdot I_i$, enriching the diversity of the training data.
To preserve the high dynamic range of the original images when converting to the sRGB color space for model training, we apply AgX tone mapping\footnote{\url{https://github.com/EaryChow/AgX}} to the sampled linear images, which has been shown to be effective in preserving more realistic highlight effects.

\section{Models}
\label{sec:models}

\subsection{Single-image Light Editing -- \paper-SV}

Our single-image editing model builds upon a pretrained text-based image editing diffusion transformer (DiT) with an architecture similar to FLUX.1 Kontext \cite{FLUX2025}.
Because full-parameter fine-tuning is prohibitively expensive, we adopt LoRA \cite{HuSWALWWC2022} for efficient lightweight adaptation.
Importantly, the base DiT model already demonstrates strong capability for general instruction-based image editing (see \cref{fig:single_image_comparison_extra}), which allows LoRA fine-tuning to specialize the model for the light editing task using only limited synthetic training data, and generalize to real-world images using the generative prior inherent in the base DiT model.
\Cref{fig:sv_model} illustrates the model design.

\begin{figure}[b]
    \centering
    \includegraphics[width=\linewidth]{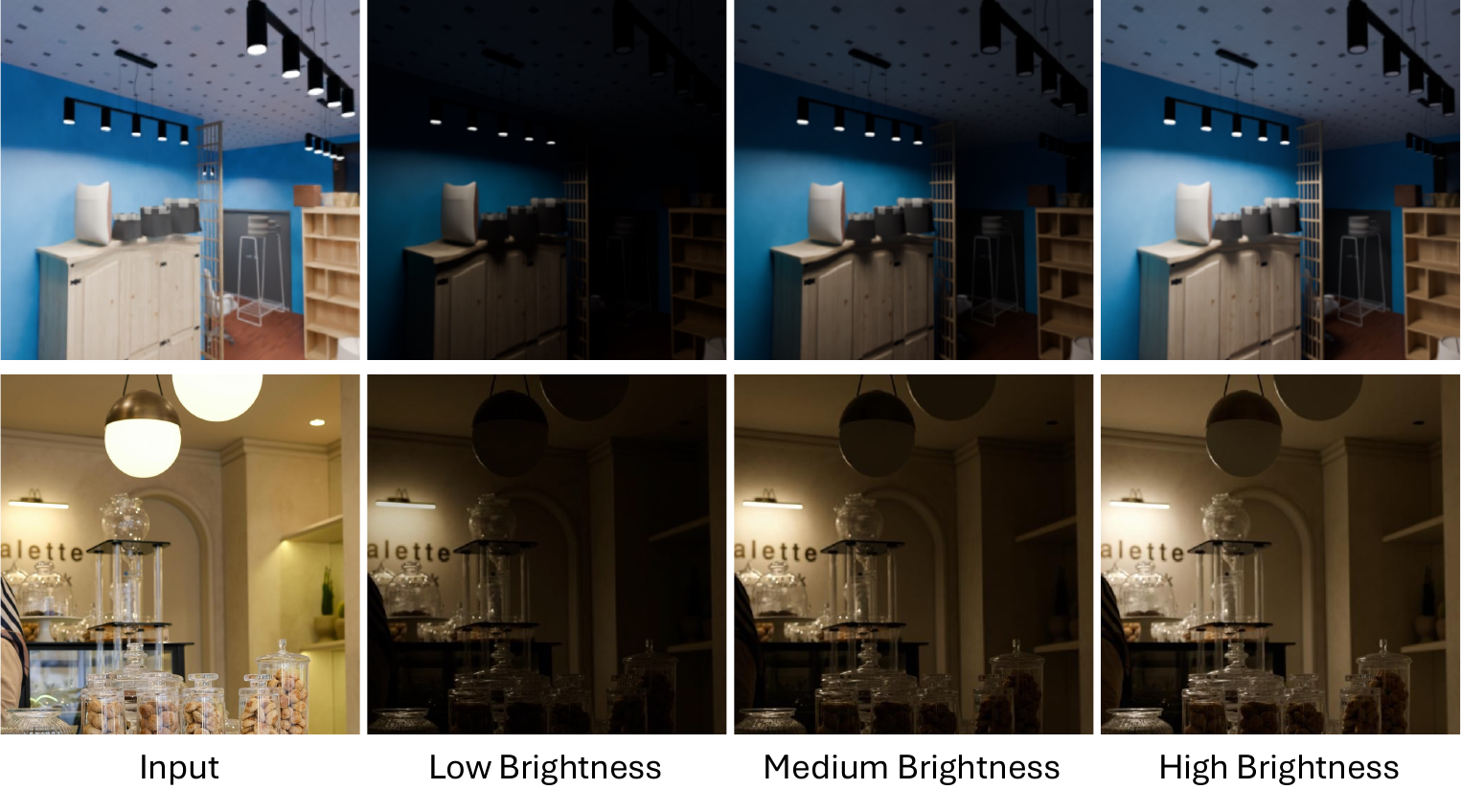}
    \caption{\label{fig:light_level_control}%
        \textbf{Brightness Level Control.} \paper-SV's OLAT decomposition at different brightness levels.
    }
\end{figure}

We use LoRA fine-tuning for two closely related light editing tasks:
(1) OLAT decomposition and (2) turning off the target light.
Below are the instruction templates we used for the two tasks:
\begin{enumerate}
\item
\emph{``\{trigger word: \texttt{OLAT}\}. Darken the room to a night scene: eliminate all light sources and window light. The only light source should be the selected \{light type\}, on at \{low,medium,high\} brightness, according to the selection mask.''}

\item
\emph{``\{trigger word: \texttt{LTOFF}\}. Only turn off the selected \{light type\}, according to the selection mask. Keep the remaining light sources unchanged.''}
\end{enumerate}
We use two distinct trigger words to differentiate between the two closely related tasks.
For the OLAT decomposition task, we further specify the desired brightness level \emph{``\{low, medium, high\} brightness''}, for the target light source.
Specifically, we instruct the model to generate LDR OLAT images corresponding to approximate exposure values of EV–4 (low), EV–2 (medium), and EV0 (high) relative to the HDR target after tone mapping.
This approach encourages the model to better learn light transport and enables reconstruction of HDR OLAT images through multi-exposure fusion \cite{DebevM1997}.
\Cref{fig:light_level_control} shows two examples of our additional brightness level control.

Other than text prompts, we also provide the light mask as a spatial prompt for the model.
The mask is processed by a single-layer MLP to match the DiT's input latent dimensions, then added to the input condition image latents.
This additional single-layer MLP is fine-tuned along with the LoRA parameters to further improve the model's ability to understand the light mask.

\input{assets/single_image_comparison_extra.tex}

\paragraph{Training details}

We insert LoRA adapters into all attention blocks of the DiT model, using a LoRA rank of 32.
Fine-tuning is performed for 3,000 iterations with a batch size of 192, leveraging the Prodigy optimizer \cite{MishcD2024} for adaptive learning rate adjustment and bias correction.
Training was conducted on 48 NVIDIA A100-40GB GPUs, completing in approximately 12 hours.

\input{assets/single_image_comparison_extra2.tex}

\subsection{Multi-view Harmonization -- \paper-MV}

Our multi-view harmonization model extends the lighting decomposition from single to multiple views.
While it is possible to use the single-view \paper-SV model for multi-view propagation (e.g., by processing diptychs or grids of images), this becomes computationally infeasible for large image sets due to high resource demands.
To address this, we use a scalable multi-view diffusion model based on a pretrained U-Net architecture (similar to CAT3D \cite{GaoHHBMSBP2024} and SEVA \cite{ZhouGVVYBTRJ2025}) for efficient multi-view lighting propagation.

The base model is pretrained to generate novel views given a few reference images.
However, our task does not require synthesizing entirely new views; instead, we have access to multi-view input images prior to lighting decomposition, and our goal is to propagate the lighting decomposition from sparse reference views to all other output views.
To accomplish this, we concatenate the original input views and the sparse light-decomposed views, together with the corresponding Plücker ray embeddings and reference view masks, as inputs to the pretrained multi-view diffusion U-Net, by extending the model's channels of the input projection layer, as illustrated in \cref{fig:mv_model}.
To distinguish between OLAT decomposition and one-light-off editing, we also include a binary mask in the input condition: all zeros for OLAT decomposition, and all ones for one-light-off editing.

\paragraph{Training details}

We conduct full-parameter fine-tuning of the U-Net for 30,000 iterations.
The training progresses in three stages: For the first 15,000 iterations, the model is trained on 4-view input batches of size 192.
In the next 10,000 iterations, the number of input views is increased to 8, with a batch size of 144.
Finally, during the last 5,000 iterations, the model is further trained with 15-view input batches of size 96.
We utilize the AdamW optimizer \cite{LoshcH2019} with a learning rate of $5 \!\times\! 10^{-5}$.
Training is performed on 48 NVIDIA A100-40GB GPUs and completes in approximately 28 hours.

\paragraph{Inference details}

Our model is fine-tuned to handle up to 15 views per forward pass.
To accommodate larger multi-view datasets, we employ a sequential multi-pass strategy.
The core idea is to process images in batches while maintaining consistency by conditioning on both the original reference views and previously generated frames.
Let $\mathcal{U}$ denote the set of unprocessed target frames and $\mathcal{P}$ the set of processed frames.
We use a distance metric $d(I_i, I_j)$ based on camera pose similarity to guide the selection process:
\begin{itemize}
    \item \textbf{Pass $1$:} We select target frames from $\mathcal{U}$ that are spatially closest to the original source reference views $\mathcal{R}^*$.
    \item \textbf{Pass $k > 1$:} We iteratively select remaining frames from $\mathcal{U}$ that are closest to any frame in the processed set $\mathcal{P}$.
\end{itemize}
For each pass, we construct a dynamic reference set $\mathcal{R}_k$ that always includes the original source references $\mathcal{R}^*$.
Additionally, if a target frame is selected due to its proximity to a previously generated frame $I_\text{prev} \in \mathcal{P}$, we optionally include $I_\text{prev}$ as a secondary reference in $\mathcal{R}_k$.
This approach allows the model to “chain” visual information from source views to distant targets, ensuring that subsequent passes remain consistent with the lighting decomposition established in earlier steps.

\subsection{Gaussian Splatting with Lighting Control}

Following the two-stage training pipeline described in \iftoggle{cvprfinal}{\cref{sec:relightable_3dgs}}{Section 3.4},
we extend the 3D Gaussian splatting model \cite{KerblKLD2023} to enable precise lighting control.
We achieve this by fitting per-light RGB parameters for each Gaussian, utilizing the light-decomposed multi-view images.

\paragraph{Stage 1: Pretraining a standard 3DGS}

We first establish the scene's geometric structure by training a standard 3D Gaussian splatting model on the original multi-view images using gsplat \cite{YeLKTYPSYHTK2025}.
Upon convergence, these pretrained Gaussians serve as the basis for incorporating additional parameters for lighting control.

\paragraph{Stage 2: Fitting per-light RGB parameters}

We augment each Gaussian from the previous stage with a set of learnable lighting parameters $\mathbf{L}_i \!\in\! \mathbb{R}^{M \times 3}$, where $M$ denotes the number of decomposed light sources (including ambient lighting).
Modeled in a linear HDR space to represent the physical accumulation of light, these parameters allow us to render the image for a specific light source $m$, denoted as $\hat{I}_m$, via standard rasterization using the Gaussians' existing geometry and the new learnable lighting parameters.
We optimize these per-light RGB parameters as follows:
\begin{enumerate}
    \item \textbf{Joint Optimization:}
    We jointly optimize the per-light RGB parameters $\mathbf{L}$ alongside the shared geometry parameters.
    This step refines the pretrained Gaussians' geometry and appearance to better align with the light-decomposed multi-view images.
    We train the Gaussians in this step for 4,000 iterations.

    \item \textbf{Light Fitting (Frozen Geometry):}
    To prevent the model from explaining lighting residuals by altering geometry, we freeze all geometric parameters and focus solely on optimizing $\mathbf{L}$ to fit the light-decomposed images for the remaining 2,000 iterations.
\end{enumerate}

\paragraph{Training Objectives}

The optimization in Stage 2 incorporates three losses: one for the fidelity of individual light renderings, one for the consistency of the recombined lighting, and one for the spatial smoothness of the Gaussian parameters.

The first term is the photometric loss between the rendered light image $\hat{I}_m$ and the ground-truth light image $I_m$:
\begin{equation}
    \mathcal{L}_\text{olat} = \frac{1}{M} \sum_{m=1}^{M} \left(\| \hat{I}_m - I_m \|_1 + \lambda \mathcal{L}_\text{D-SSIM}(\hat{I}_m, I_m) \right) \text{.}
\end{equation}

The second term is a composition consistency loss between the recombined lighting $\hat{I}_\text{comp} = \sum_{m} w_m \hat{I}_m$ and the original input image $I_\text{ori}$, where $w_m$ is a learnable per-light scaling factor for light recombination.
This ensures that the learned lighting coefficients, when recombined, accurately reconstruct the original appearance:
\begin{equation}
    \mathcal{L}_\text{comp} = \| \mathcal{T}(\hat{I}_\text{comp}) - I_\text{ori} \|_1 \text{,}
\end{equation}
where $\mathcal{T}(x) = (x + \beta)^{\frac{1}{\gamma}}$ is a differentiable tone mapping function with a learnable gamma $\gamma$ and offset $\beta$.

The third component is a spatial smoothness loss.
To mitigate high-frequency noise where adjacent Gaussians learn divergent light responses, we impose a spatial smoothness regularizer based on K-Nearest Neighbors (KNN).
For each Gaussian $i$, we penalize the deviation of its lighting coefficients from its $K$ nearest spatial neighbors $\mathcal{N}(i)$:
\begin{equation}
    \mathcal{L}_\text{smooth} = \frac{1}{NK} \sum_{i=1}^{N} \sum_{j \in \mathcal{N}(i)} \| \mathbf{L}_i - \mathbf{L}_j \|_2^2 \text{.}
\end{equation}
This encourages local consistency in light reflectance, helping to reduce sparkling artifacts during relighting.
We apply this loss every 100 optimization iterations after the initial 4,000 iterations.

The final objective is the weighted sum of three losses:
\begin{equation}
    \mathcal{L} = \mathcal{L}_\text{olat} + \lambda_\text{comp} \mathcal{L}_\text{comp} + \lambda_\text{smooth} \mathcal{L}_\text{smooth} \text{.}
\end{equation}

\begin{figure*}[t]
    \centering
    \begin{tabular}{@{}c@{\hspace{4pt}}c@{}}
        \begin{minipage}[c]{0.11\linewidth}
            \raggedleft
            {\footnotesize Input} \\[5.5em]
            {\footnotesize LuxRemix-SV} \\[5.5em]
            {\footnotesize LuxRemix-MV}
        \end{minipage}
        &
        \begin{minipage}[c]{0.83\linewidth}
            \begin{tikzpicture}[font=\sffamily\footnotesize]
                \node[inner sep=0pt] (img) at (0,0) {%
                    \includegraphics[width=\linewidth]{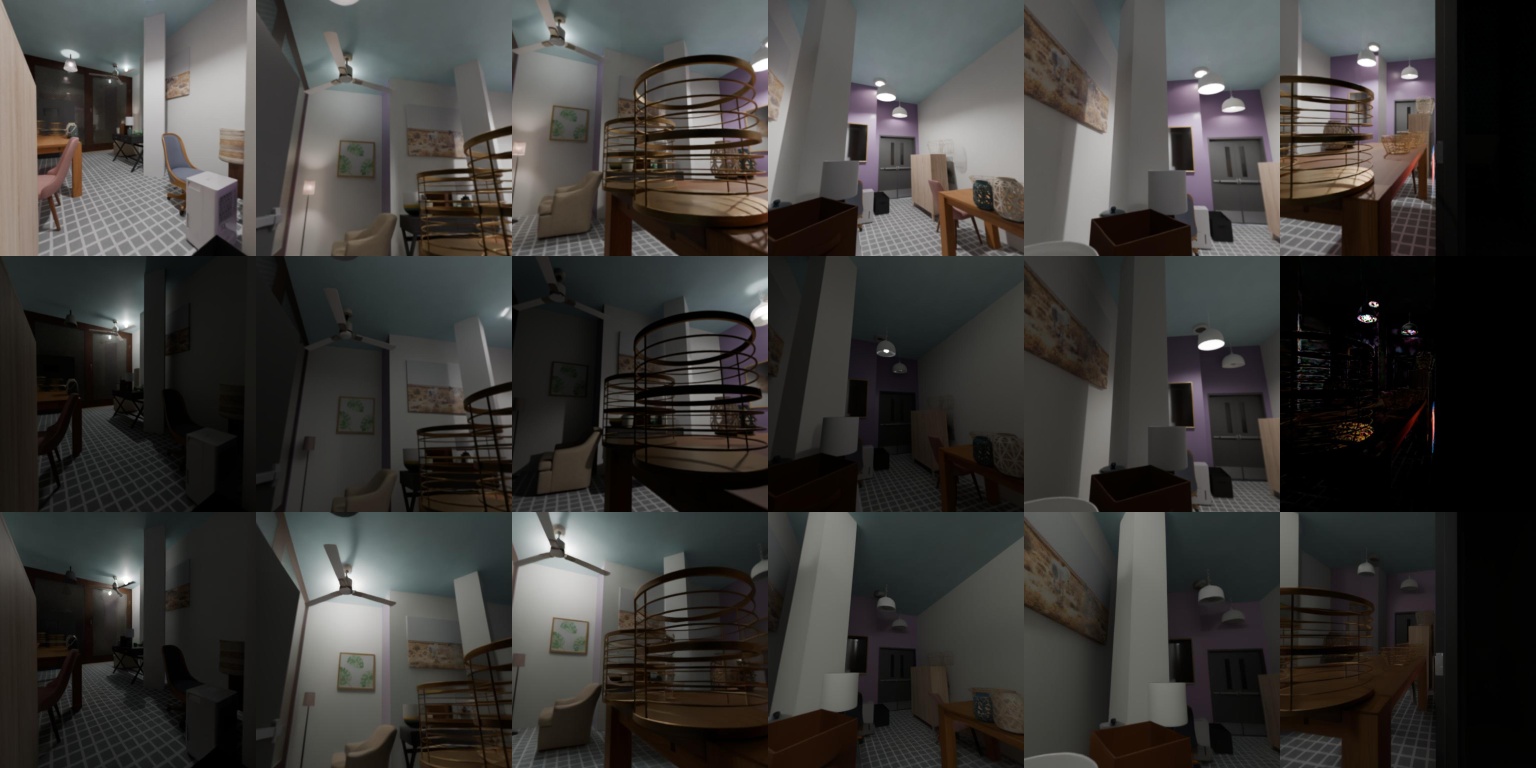}%
                };

                \draw[stealth-stealth, red, line width=0.7mm]
                    ([xshift=38pt, yshift=-88pt]img.north west) --
                    ([xshift=87pt, yshift=-88pt]img.north west);

                \draw[-stealth, red, line width=0.7mm] ([xshift=149pt, yshift=-112pt]img.north west) -- ++(0pt, 26pt);

                \draw[-stealth, red, line width=0.7mm] ([xshift=105pt, yshift=-84pt]img.north west) -- ++(26pt, 0pt);
                \draw[-stealth, red, line width=0.7mm] ([xshift=173pt, yshift=-84pt]img.north west) -- ++(26pt, 0pt);
                \draw[-stealth, red, line width=0.7mm] ([xshift=398pt, yshift=-84pt]img.north west) -- ++(-26pt, 0pt);

                \draw[stealth-stealth, red, line width=0.7mm] ([xshift=245pt, yshift=-94pt]img.north west) -- ([xshift=318pt, yshift=-93pt]img.north west);
            \end{tikzpicture}
        \end{minipage}
    \end{tabular}
    \caption{\textbf{Single-View vs.\ Multi-View Consistency.} Lighting edits on a synthetic test scene.}
    \label{fig:sv_mv_consistency_comparison}
\end{figure*}

After training, we can randomly blend multiple per-light renderings from the same set of Gaussians to get consistent relighting results.
We include screen recording videos in the supplement to show our interactive relighting with 3DGS.

\section{Additional Results and Comparisons}
\label{sec:additional_results}

\subsection{Single-Image Lighting Editing}

We show additional editing results in \cref{fig:single_image_comparison_extra,fig:single_image_comparison_extra2}.
In \cref{fig:single_image_comparison_extra}, we show results of switching lights off and on, with various light colors.
The baseline comparisons to FLUX.1 Kontext \cite{FLUX2025}, Qwen-Image \cite{WuLZLGYYBXCCTZWYYCLLZMWNCCPQWWYWFXWZZWCL2025} and Gemini 2.5 Flash Image a.k.a. Nano Banana show prompt-based editing results trying to match our results.
\Cref{fig:single_image_comparison_extra2} shows more advanced editing results with two editable light sources and a mixture of out-of-distribution light shapes and novel colors.
In the supplemental material, we include a video demonstrating the decomposition of seven distinct light sources, and recombining them in many different ways.

\subsection{Multi-view Lighting Harmonization}

\Cref{fig:lighting_harmonization_extra} shows additional multi-view lighting harmonization results.
Note how our single-view model \paper-SV can recover individual OLAT lights with plausible shadows for the plant in the top scene.
Our multi-view harmonization model \paper-MV then propagates these single-view lighting decompositions consistently across views.
For the middle scene, our model can cleanly disentangle the three hanging lights.
And for the bottom scene, our model can switch on the standing lamp in the center of the scene (see “OLAT 3”).
Note also that the ambient lighting in this scene correctly includes the out-of-view ceiling lights.

\subsection{Real-World Paired Evaluation on LightLab}
\label{sec:lightlab_eval_supp}

To complement our synthetic evaluation, we include additional real-world paired benchmarking on the LightLab examples.
For each method, we tune light editing parameters to best match each paired target image and report photometric metrics.
\Cref{tab:lightlab_comparison_supp} summarizes results on all 10 available paired examples.
While LightLab achieves the strongest scores on this real-data benchmark, our method remains competitive despite being trained on synthetic data only.

\begin{table}[t]
    \caption{\textbf{Real-World Evaluation on LightLab.} Quantitative comparison on 10 paired real-world examples.}
    \label{tab:lightlab_comparison_supp}
    \centering
    \small
    \begin{tabular}{lccc}
    \toprule
    Method & PSNR $\uparrow$ & SSIM $\uparrow$ & LPIPS $\downarrow$ \\
    \midrule
    ScribbleLight \cite{ChoiWPBS2025} & 15.04 & 0.465 & 0.478 \\
    Qwen-Image \cite{WuLZLGYYBXCCTZWYYCLLZMWNCCPQWWYWFXWZZWCL2025} & 19.20 & 0.735 & 0.196 \\
    LightLab \cite{MagarHTPRSH2025} & 21.80 & 0.791 & 0.144 \\[3pt]
    Ours & 20.93 & 0.771 & 0.168 \\
    \bottomrule
    \end{tabular}
    \vspace{-4pt}
\end{table}

\subsection{Single-View versus Multi-View Consistency}
\label{sec:sv_mv_consistency_supp}

We further compare per-frame single-view editing against our multi-view harmonization, as shown in \cref{fig:sv_mv_consistency_comparison}.
Applying \paper-SV independently to each view can introduce cross-view inconsistencies in shadows and lighting appearance.
In contrast, \paper-MV propagates decomposition jointly across views and yields more consistent multi-view results.

\input{assets/failure_cases.tex}

\subsection{Failure cases}

\Cref{fig:failure_cases} shows some failure cases of our single-image light decomposition and editing method.
In our synthetic training data, the shape of the light spread is biased towards conical shapes, which does not always match real-world lights.
In multi-light images, our lighting decomposition sometimes fails to adhere to the provided light mask.
Using a different random seed can result in different OLAT decompositions.

\input{assets/lighting_harmonization_extra.tex}

%% file: assets/single_image_comparison_extra.tex
\begin{figure*}[p]
  \centering
  \begin{tikzpicture}[font=\sffamily\footnotesize]
    \def\imgwidth{0.14\linewidth}
    \def\imgsep{2pt}
    
    \def\scene{jean-philippe-delberghe}  %
    \node[inner sep=0pt, anchor=north west] (img1-input) at (0,0)
    {\includegraphics[width=\imgwidth]{assets/single_image_comparison/pexels/\scene/input.jpg}};
    \node[inner sep=0pt, anchor=north west] (img1-mask) at ([xshift=\imgsep]img1-input.north east) 
    {\includegraphics[width=\imgwidth]{assets/single_image_comparison/pexels/\scene/mask_0.jpg}};
    \node[inner sep=0pt, anchor=north west] (img1-flux) at ([xshift=\imgsep]img1-mask.north east)
    {\includegraphics[width=\imgwidth]{assets/single_image_comparison/pexels/\scene/flux.jpg}};
    \node[inner sep=0pt, anchor=north west] (img1-qwen) at ([xshift=\imgsep]img1-flux.north east)
    {\includegraphics[width=\imgwidth]{assets/single_image_comparison/pexels/\scene/qwen.jpg}};
    \node[inner sep=0pt, anchor=north west] (img1-nb)   at ([xshift=\imgsep]img1-qwen.north east)
    {\includegraphics[width=\imgwidth]{assets/single_image_comparison/pexels/\scene/nb.jpg}};
    \node[inner sep=0pt, anchor=north west] (img1-ours) at ([xshift=\imgsep]img1-nb.north east)
    {\includegraphics[width=\imgwidth]{assets/single_image_comparison/pexels/\scene/ours.jpg}};
    
    \def\scene{ilya-shakir-1278798-2440471}  %
    \node[inner sep=0pt, anchor=north, below=\imgsep] (img2-input) at (img1-input.south)
    {\includegraphics[width=\imgwidth]{assets/single_image_comparison/pexels/\scene/input.jpg}};
    \node[inner sep=0pt, anchor=north west] (img2-mask) at ([xshift=\imgsep]img2-input.north east) 
    {\includegraphics[width=\imgwidth]{assets/single_image_comparison/pexels/\scene/mask_0.jpg}};
    \node[inner sep=0pt, anchor=north west] (img2-flux) at ([xshift=\imgsep]img2-mask.north east)
    {\includegraphics[width=\imgwidth]{assets/single_image_comparison/pexels/\scene/flux.jpg}};
    \node[inner sep=0pt, anchor=north west] (img2-qwen) at ([xshift=\imgsep]img2-flux.north east)
    {\includegraphics[width=\imgwidth]{assets/single_image_comparison/pexels/\scene/qwen.jpg}};
    \node[inner sep=0pt, anchor=north west] (img2-nb)   at ([xshift=\imgsep]img2-qwen.north east)
    {\includegraphics[width=\imgwidth]{assets/single_image_comparison/pexels/\scene/nb.jpg}};
    \node[inner sep=0pt, anchor=north west] (img2-ours) at ([xshift=\imgsep]img2-nb.north east)
    {\includegraphics[width=\imgwidth]{assets/single_image_comparison/pexels/\scene/ours.jpg}};
    
    \def\scene{pixabay-265129}
    \node[inner sep=0pt, anchor=north, below=\imgsep] (img3-input) at (img2-input.south)
    {\includegraphics[width=\imgwidth]{assets/single_image_comparison/pexels/\scene/input.jpg}};
    \node[inner sep=0pt, anchor=north west] (img3-mask) at ([xshift=\imgsep]img3-input.north east) 
    {\includegraphics[width=\imgwidth]{assets/single_image_comparison/pexels/\scene/mask_0.jpg}};
    \node[inner sep=0pt, anchor=north west] (img3-flux) at ([xshift=\imgsep]img3-mask.north east)
    {\includegraphics[width=\imgwidth]{assets/single_image_comparison/pexels/\scene/flux.jpg}};
    \node[inner sep=0pt, anchor=north west] (img3-qwen) at ([xshift=\imgsep]img3-flux.north east)
    {\includegraphics[width=\imgwidth]{assets/single_image_comparison/pexels/\scene/qwen.jpg}};
    \node[inner sep=0pt, anchor=north west] (img3-nb)   at ([xshift=\imgsep]img3-qwen.north east)
    {\includegraphics[width=\imgwidth]{assets/single_image_comparison/pexels/\scene/nb.jpg}};
    \node[inner sep=0pt, anchor=north west] (img3-ours) at ([xshift=\imgsep]img3-nb.north east)
    {\includegraphics[width=\imgwidth]{assets/single_image_comparison/pexels/\scene/ours.jpg}};
    
    \def\scene{polina-kovaleva-5546811}
    \node[inner sep=0pt, anchor=north, below=\imgsep] (img4-input) at (img3-input.south)
    {\includegraphics[width=\imgwidth]{assets/single_image_comparison/pexels/\scene/input.jpg}};
    \node[inner sep=0pt, anchor=north west] (img4-mask) at ([xshift=\imgsep]img4-input.north east) 
    {\includegraphics[width=\imgwidth]{assets/single_image_comparison/pexels/\scene/mask_0.jpg}};
    \node[inner sep=0pt, anchor=north west] (img4-flux) at ([xshift=\imgsep]img4-mask.north east)
    {\includegraphics[width=\imgwidth]{assets/single_image_comparison/pexels/\scene/flux.jpg}};
    \node[inner sep=0pt, anchor=north west] (img4-qwen) at ([xshift=\imgsep]img4-flux.north east)
    {\includegraphics[width=\imgwidth]{assets/single_image_comparison/pexels/\scene/qwen.jpg}};
    \node[inner sep=0pt, anchor=north west] (img4-nb)   at ([xshift=\imgsep]img4-qwen.north east)
    {\includegraphics[width=\imgwidth]{assets/single_image_comparison/pexels/\scene/nb.jpg}};
    \node[inner sep=0pt, anchor=north west] (img4-ours) at ([xshift=\imgsep]img4-nb.north east)
    {\includegraphics[width=\imgwidth]{assets/single_image_comparison/pexels/\scene/ours.jpg}};
    
    \def\scene{burst-545034}
    \node[inner sep=0pt, anchor=north, below=\imgsep] (img5-input) at (img4-input.south)
    {\includegraphics[width=\imgwidth]{assets/single_image_comparison/pexels/\scene/input.jpg}};
    \node[inner sep=0pt, anchor=north west] (img5-mask) at ([xshift=\imgsep]img5-input.north east) 
    {\includegraphics[width=\imgwidth]{assets/single_image_comparison/pexels/\scene/mask_0.jpg}};
    \node[inner sep=0pt, anchor=north west] (img5-flux) at ([xshift=\imgsep]img5-mask.north east)
    {\includegraphics[width=\imgwidth]{assets/single_image_comparison/pexels/\scene/flux.jpg}};
    \node[inner sep=0pt, anchor=north west] (img5-qwen) at ([xshift=\imgsep]img5-flux.north east)
    {\includegraphics[width=\imgwidth]{assets/single_image_comparison/pexels/\scene/qwen.jpg}};
    \node[inner sep=0pt, anchor=north west] (img5-nb)   at ([xshift=\imgsep]img5-qwen.north east)
    {\includegraphics[width=\imgwidth]{assets/single_image_comparison/pexels/\scene/nb.jpg}};
    \node[inner sep=0pt, anchor=north west] (img5-ours) at ([xshift=\imgsep]img5-nb.north east)
    {\includegraphics[width=\imgwidth]{assets/single_image_comparison/pexels/\scene/ours.jpg}};

    \def\scene{mikhail-nilov-6981121}
    \node[inner sep=0pt, anchor=north, below=\imgsep] (img6-input) at (img5-input.south)
    {\includegraphics[width=\imgwidth]{assets/single_image_comparison/pexels/\scene/input.jpg}};
    \node[inner sep=0pt, anchor=north west] (img6-mask) at ([xshift=\imgsep]img6-input.north east) 
    {\includegraphics[width=\imgwidth]{assets/single_image_comparison/pexels/\scene/mask_0.jpg}};
    \node[inner sep=0pt, anchor=north west] (img6-flux) at ([xshift=\imgsep]img6-mask.north east)
    {\includegraphics[width=\imgwidth]{assets/single_image_comparison/pexels/\scene/flux.jpg}};
    \node[inner sep=0pt, anchor=north west] (img6-qwen) at ([xshift=\imgsep]img6-flux.north east)
    {\includegraphics[width=\imgwidth]{assets/single_image_comparison/pexels/\scene/qwen.jpg}};
    \node[inner sep=0pt, anchor=north west] (img6-nb)   at ([xshift=\imgsep]img6-qwen.north east)
    {\includegraphics[width=\imgwidth]{assets/single_image_comparison/pexels/\scene/nb.jpg}};
    \node[inner sep=0pt, anchor=north west] (img6-ours) at ([xshift=\imgsep]img6-nb.north east)
    {\includegraphics[width=\imgwidth]{assets/single_image_comparison/pexels/\scene/ours.jpg}};

    \def\scene{marina-podrez-3269296-11673566}
    \node[inner sep=0pt, anchor=north, below=\imgsep] (img7-input) at (img6-input.south)
    {\includegraphics[width=\imgwidth]{assets/single_image_comparison/pexels/\scene/input.jpg}};
    \node[inner sep=0pt, anchor=north west] (img7-mask) at ([xshift=\imgsep]img7-input.north east) 
    {\includegraphics[width=\imgwidth]{assets/single_image_comparison/pexels/\scene/mask_0.jpg}};
    \node[inner sep=0pt, anchor=north west] (img7-flux) at ([xshift=\imgsep]img7-mask.north east)
    {\includegraphics[width=\imgwidth]{assets/single_image_comparison/pexels/\scene/flux.jpg}};
    \node[inner sep=0pt, anchor=north west] (img7-qwen) at ([xshift=\imgsep]img7-flux.north east)
    {\includegraphics[width=\imgwidth]{assets/single_image_comparison/pexels/\scene/qwen.jpg}};
    \node[inner sep=0pt, anchor=north west] (img7-nb)   at ([xshift=\imgsep]img7-qwen.north east)
    {\includegraphics[width=\imgwidth]{assets/single_image_comparison/pexels/\scene/nb.jpg}};
    \node[inner sep=0pt, anchor=north west] (img7-ours) at ([xshift=\imgsep]img7-nb.north east)
    {\includegraphics[width=\imgwidth]{assets/single_image_comparison/pexels/\scene/ours.jpg}};

    \def\scene{becca-tapert-p6h5U}
    \node[inner sep=0pt, anchor=north, below=\imgsep] (img8-input) at (img7-input.south)
    {\includegraphics[width=\imgwidth]{assets/single_image_comparison/pexels/\scene/input.jpg}};
    \node[inner sep=0pt, anchor=north west] (img8-mask) at ([xshift=\imgsep]img8-input.north east) 
    {\includegraphics[width=\imgwidth]{assets/single_image_comparison/pexels/\scene/mask_0.jpg}};
    \node[inner sep=0pt, anchor=north west] (img8-flux) at ([xshift=\imgsep]img8-mask.north east)
    {\includegraphics[width=\imgwidth]{assets/single_image_comparison/pexels/\scene/flux.jpg}};
    \node[inner sep=0pt, anchor=north west] (img8-qwen) at ([xshift=\imgsep]img8-flux.north east)
    {\includegraphics[width=\imgwidth]{assets/single_image_comparison/pexels/\scene/qwen.jpg}};
    \node[inner sep=0pt, anchor=north west] (img8-nb)   at ([xshift=\imgsep]img8-qwen.north east)
    {\includegraphics[width=\imgwidth]{assets/single_image_comparison/pexels/\scene/nb.jpg}};
    \node[inner sep=0pt, anchor=north west] (img8-ours) at ([xshift=\imgsep]img8-nb.north east)
    {\includegraphics[width=\imgwidth]{assets/single_image_comparison/pexels/\scene/ours.jpg}};

    \node[anchor=south, text depth=0.25ex, inner sep=2pt] at ([yshift=2pt]img1-input.north) {Input Image};
    \node[anchor=south, text depth=0.25ex, inner sep=2pt] at ([yshift=2pt]img1-mask.north) {Light Mask};
    \node[anchor=south, text depth=0.25ex, inner sep=2pt] at ([yshift=2pt]img1-flux.north) {FLUX.1 Kontext \cite{FLUX2025}};
    \node[anchor=south, text depth=0.25ex, inner sep=2pt] at ([yshift=2pt]img1-qwen.north) {Qwen-Image \cite{WuLZLGYYBXCCTZWYYCLLZMWNCCPQWWYWFXWZZWCL2025}};
    \node[anchor=south, text depth=0.25ex, inner sep=2pt] at ([yshift=2pt]img1-nb.north) {Nano Banana};
    \node[anchor=south, text depth=0.25ex, inner sep=2pt] at ([yshift=2pt]img1-ours.north) {\paper-SV (ours)};
    
  \end{tikzpicture}
  \caption{\label{fig:single_image_comparison_extra}%
    \textbf{Additional Comparisons of Single-Image Light Editing.}
    Given an input image and a light mask (left), we compare our light editing results with baseline methods for switching lights on and off.
    FLUX.1 Kontext \cite{FLUX2025} sometimes removes or modifies image details.
    Qwen-Image \cite{WuLZLGYYBXCCTZWYYCLLZMWNCCPQWWYWFXWZZWCL2025} sometimes removes lights entirely.
    Nano Banana produces plausible light edits.
    Our \paper-SV model produces convincing results throughout with fine-grained controllability over individual light intensities and colors.
    Real-world images from Pexels.
  }
\end{figure*}

%% file: assets/single_image_comparison_extra2.tex
\begin{figure*}[t]
  \centering
  \begin{tikzpicture}[font=\sffamily\footnotesize]
    \def\imgwidth{0.162\linewidth}
    \def\imgsep{2pt}
    
    \def\scene{introspectivedsgn-27703394}
    \node[inner sep=0pt, anchor=north west] (img1-input) at (0,0)
    {\includegraphics[width=\imgwidth]{assets/single_image_comparison/pexels/\scene/input.jpg}};
    \node[inner sep=0pt, anchor=north west] (img1-mask0) at ([xshift=\imgsep]img1-input.north east) {\includegraphics[width=\imgwidth]{assets/single_image_comparison/pexels/\scene/mask_0.jpg}};
    \node[inner sep=0pt, anchor=north west] (img1-mask1) at ([xshift=\imgsep]img1-mask0.north east) {\includegraphics[width=\imgwidth]{assets/single_image_comparison/pexels/\scene/mask_1.jpg}};
    \node[inner sep=0pt, anchor=north west] (img1-qwen)  at ([xshift=\imgsep]img1-mask1.north east)
    {\includegraphics[width=\imgwidth]{assets/single_image_comparison/pexels/\scene/qwen.jpg}};
    \node[inner sep=0pt, anchor=north west] (img1-nb)    at ([xshift=\imgsep]img1-qwen.north east)
    {\includegraphics[width=\imgwidth]{assets/single_image_comparison/pexels/\scene/nb.jpg}};
    \node[inner sep=0pt, anchor=north west] (img1-ours)  at ([xshift=\imgsep]img1-nb.north east)
    {\includegraphics[width=\imgwidth]{assets/single_image_comparison/pexels/\scene/ours.jpg}};
    
    \def\scene{juanpphotoandvideo-967016}
    \node[inner sep=0pt, anchor=north, below=\imgsep] (img2-input) at (img1-input.south)
    {\includegraphics[width=\imgwidth]{assets/single_image_comparison/pexels/\scene/input.jpg}};
    \node[inner sep=0pt, anchor=north west] (img2-mask0) at ([xshift=\imgsep]img2-input.north east) {\includegraphics[width=\imgwidth]{assets/single_image_comparison/pexels/\scene/mask_0.jpg}};
    \node[inner sep=0pt, anchor=north west] (img2-mask1) at ([xshift=\imgsep]img2-mask0.north east) {\includegraphics[width=\imgwidth]{assets/single_image_comparison/pexels/\scene/mask_1.jpg}};
    \node[inner sep=0pt, anchor=north west] (img2-qwen)  at ([xshift=\imgsep]img2-mask1.north east)
    {\includegraphics[width=\imgwidth]{assets/single_image_comparison/pexels/\scene/qwen.jpg}};
    \node[inner sep=0pt, anchor=north west] (img2-nb)    at ([xshift=\imgsep]img2-qwen.north east)
    {\includegraphics[width=\imgwidth]{assets/single_image_comparison/pexels/\scene/nb.jpg}};
    \node[inner sep=0pt, anchor=north west] (img2-ours)  at ([xshift=\imgsep]img2-nb.north east)
    {\includegraphics[width=\imgwidth]{assets/single_image_comparison/pexels/\scene/ours.jpg}};

    \def\scene{z1_christian-mackie-cc0Gg3BegjE-unsplash}
    \node[inner sep=0pt, anchor=north, below=\imgsep] (img3-input) at (img2-input.south)
    {\includegraphics[width=\imgwidth]{assets/single_image_comparison/pexels/\scene/input.jpg}};
    \node[inner sep=0pt, anchor=north west] (img3-mask0) at ([xshift=\imgsep]img3-input.north east) {\includegraphics[width=\imgwidth]{assets/single_image_comparison/pexels/\scene/mask_0.jpg}};
    \node[inner sep=0pt, anchor=north west] (img3-mask1) at ([xshift=\imgsep]img3-mask0.north east) {\includegraphics[width=\imgwidth]{assets/single_image_comparison/pexels/\scene/mask_1.jpg}};
    \node[inner sep=0pt, anchor=north west] (img3-qwen)  at ([xshift=\imgsep]img3-mask1.north east)
    {\includegraphics[width=\imgwidth]{assets/single_image_comparison/pexels/\scene/qwen.jpg}};
    \node[inner sep=0pt, anchor=north west] (img3-nb)    at ([xshift=\imgsep]img3-qwen.north east)
    {\includegraphics[width=\imgwidth]{assets/single_image_comparison/pexels/\scene/nb.jpg}};
    \node[inner sep=0pt, anchor=north west] (img3-ours)  at ([xshift=\imgsep]img3-nb.north east)
    {\includegraphics[width=\imgwidth]{assets/single_image_comparison/pexels/\scene/ours.jpg}};

    \def\scene{z_mminschlxxx-31827772}    
    \node[inner sep=0pt, anchor=north, below=\imgsep] (img4-input) at (img3-input.south)
    {\includegraphics[width=\imgwidth]{assets/single_image_comparison/pexels/\scene/input.jpg}};
    \node[inner sep=0pt, anchor=north west] (img4-mask0) at ([xshift=\imgsep]img4-input.north east) {\includegraphics[width=\imgwidth]{assets/single_image_comparison/pexels/\scene/mask_0.jpg}};
    \node[inner sep=0pt, anchor=north west] (img4-mask1) at ([xshift=\imgsep]img4-mask0.north east) {\includegraphics[width=\imgwidth]{assets/single_image_comparison/pexels/\scene/mask_1.jpg}};
    \node[inner sep=0pt, anchor=north west] (img4-qwen)  at ([xshift=\imgsep]img4-mask1.north east)
    {\includegraphics[width=\imgwidth]{assets/single_image_comparison/pexels/\scene/qwen.jpg}};
    \node[inner sep=0pt, anchor=north west] (img4-nb)    at ([xshift=\imgsep]img4-qwen.north east)
    {\includegraphics[width=\imgwidth]{assets/single_image_comparison/pexels/\scene/nb.jpg}};
    \node[inner sep=0pt, anchor=north west] (img4-ours)  at ([xshift=\imgsep]img4-nb.north east)
    {\includegraphics[width=\imgwidth]{assets/single_image_comparison/pexels/\scene/ours.jpg}};

    \node[anchor=south, text depth=0.25ex, inner sep=2pt] at ([yshift=2pt]img1-input.north) {Input Image};
    \node[anchor=south, text depth=0.25ex, inner sep=2pt] at ([yshift=2pt]img1-mask0.north) {Light Mask 1};
    \node[anchor=south, text depth=0.25ex, inner sep=2pt] at ([yshift=2pt]img1-mask1.north) {Light Mask 2};
    \node[anchor=south, text depth=0.25ex, inner sep=2pt] at ([yshift=2pt]img1-qwen.north) {Qwen-Image \cite{WuLZLGYYBXCCTZWYYCLLZMWNCCPQWWYWFXWZZWCL2025}};
    \node[anchor=south, text depth=0.25ex, inner sep=2pt] at ([yshift=2pt]img1-nb.north) {Nano Banana};
    \node[anchor=south, text depth=0.25ex, inner sep=2pt] at ([yshift=2pt]img1-ours.north) {\paper-SV (ours)};
    
  \end{tikzpicture}
  \caption{\label{fig:single_image_comparison_extra2}%
    \textbf{Comparisons of Single-Image Light Editing with Multiple Masks.}
    Given an input image and two light masks (left), we compare our light editing results (right) with baseline image editing models.
    While Qwen-Image \cite{WuLZLGYYBXCCTZWYYCLLZMWNCCPQWWYWFXWZZWCL2025} and Nano Banana enable basic prompt-based light editing, only our \paper-SV approach enables realistic and fine-grained light editing control via light mask conditioning.
    Real-world images from Pexels and Unsplash.
  }
\end{figure*}

%% file: assets/failure_cases.tex
\begin{figure}[t]
  \centering
  \begin{tikzpicture}[
    label/.style={anchor=north west, fill=black, fill opacity=0.6, text=white, text opacity=1, inner sep=2pt, font=\sffamily\footnotesize}
  ]
    \def\imgwidth{0.325\linewidth}
    \def\colsep{2pt}
    \def\rowsep{2pt}

    \def\scene{014658}
    
    \node[inner sep=0pt, anchor=north west] (img11) at (0,0) {%
      \includegraphics[width=\imgwidth]{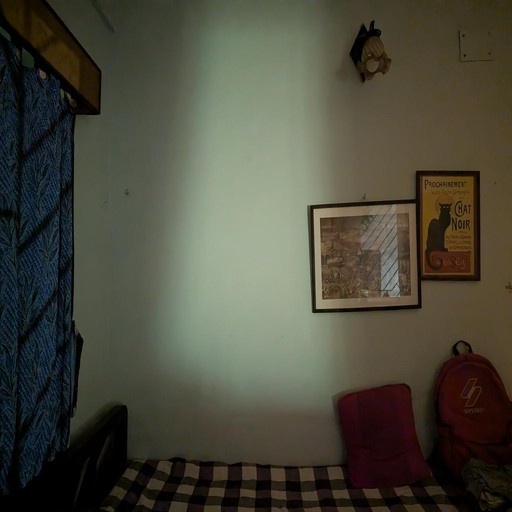}%
    };
    \node[inner sep=0pt, anchor=north west] (img12) at ([xshift=\colsep]img11.north east) {%
      \includegraphics[width=\imgwidth]{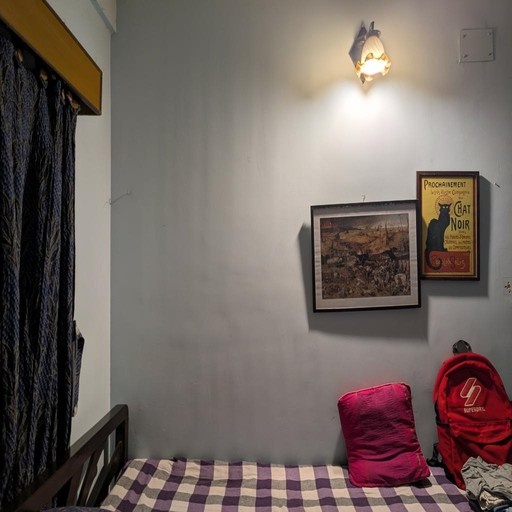}%
    };
    \node[inner sep=0pt, anchor=north west] (img13) at ([xshift=\colsep]img12.north east) {%
      \includegraphics[width=\imgwidth]{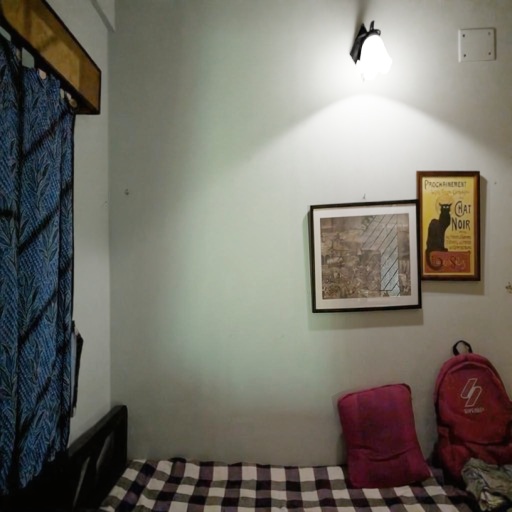}%
    };
    \draw[-stealth, red, line width=1.5mm] ([xshift=20pt, yshift=-20pt]img13.north west) -- ([xshift=40pt, yshift=-20pt]img13.north west);

    \node[inner sep=0pt, anchor=north west] (img21) at ([yshift=-\rowsep]img11.south west) {%
    \includegraphics[width=\imgwidth]{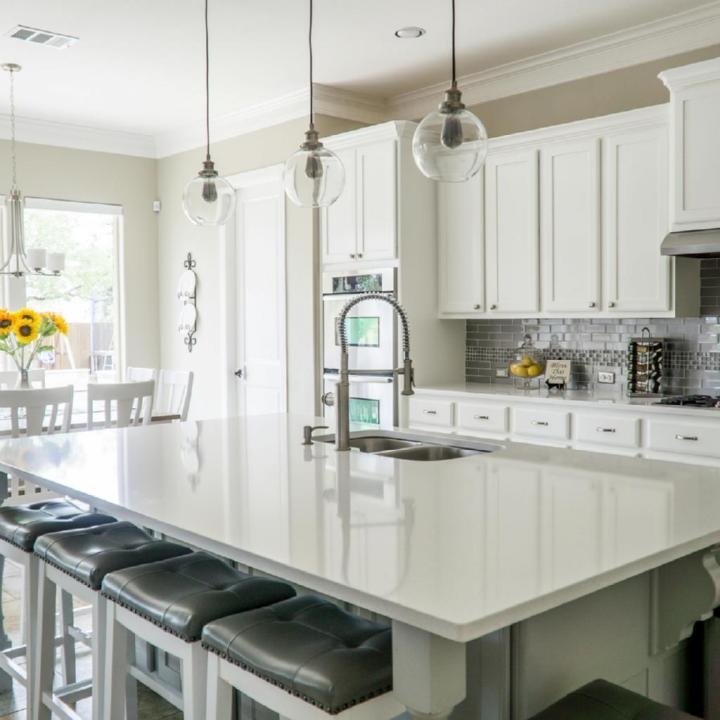}%
  };
  \node[inner sep=0pt, anchor=north west] (img22) at ([xshift=\colsep]img21.north east) {%
    \includegraphics[width=\imgwidth]{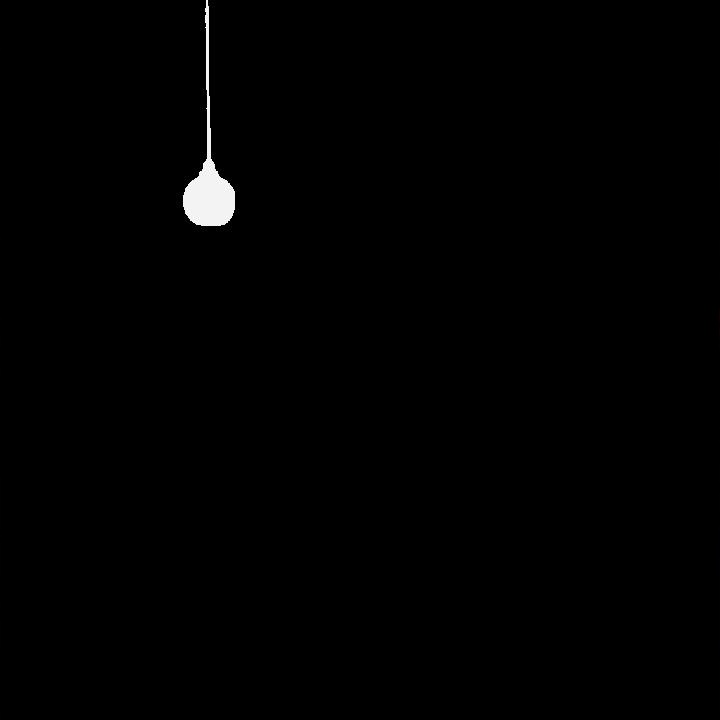}%
  };
  \node[inner sep=0pt, anchor=north west] (img23) at ([xshift=\colsep]img22.north east) {%
    \includegraphics[width=\imgwidth]{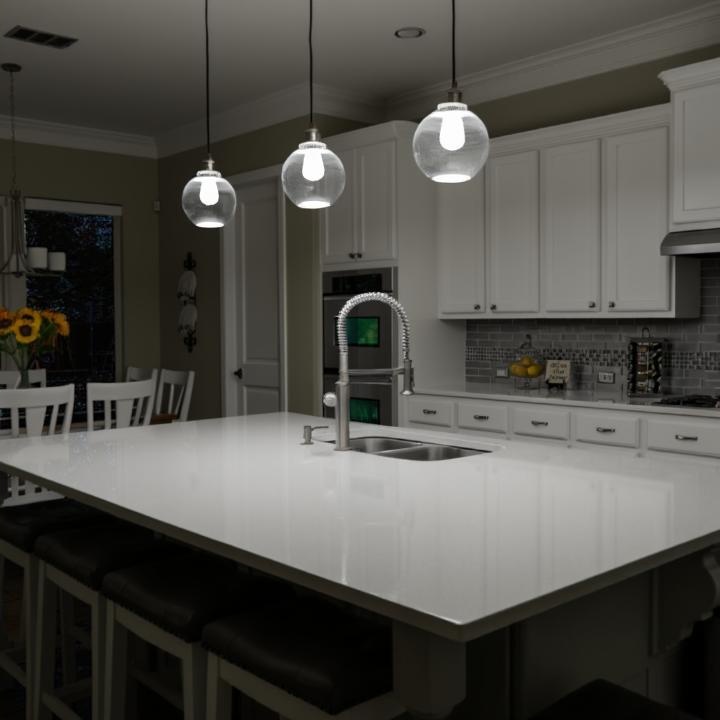}%
  };
  \draw[-stealth, red, line width=1.5mm] ([xshift=34.5pt, yshift=-45pt]img23.north west) -- ([xshift=34.5pt, yshift=-25pt]img23.north west);
  \draw[-stealth, red, line width=1.5mm] ([xshift=49pt, yshift=-42pt]img23.north west) -- ([xshift=49pt, yshift=-22pt]img23.north west);
    
  \node[inner sep=0pt, anchor=north west] (img31) at ([yshift=-\rowsep]img21.south west) {%
    \includegraphics[width=\imgwidth]{assets/failure_cases/falling4utah-1080721/input.jpg}%
  };
  \node[inner sep=0pt, anchor=north west] (img32) at ([xshift=\colsep]img31.north east) {%
    \includegraphics[width=\imgwidth]{assets/failure_cases/falling4utah-1080721/mask02.jpg}%
  };
  \node[inner sep=0pt, anchor=north west] (img33) at ([xshift=\colsep]img32.north east) {%
    \includegraphics[width=\imgwidth]{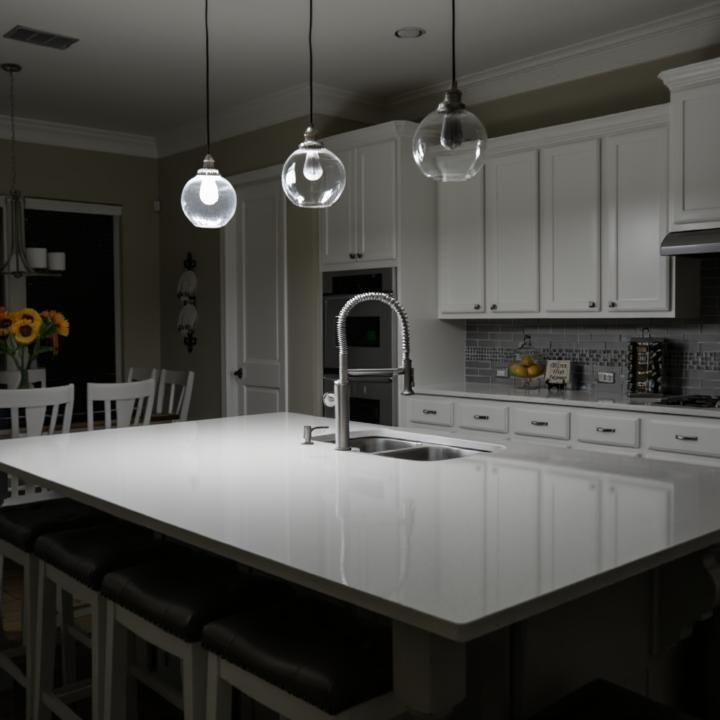}%
  };

    \node[label] at (img11.north west) {Input image};
    \node[label] at (img12.north west) {Ground truth};
    \node[label] at (img13.north west) {Our relighting result};
    \node[label] at (img21.north west) {Input image};
    \node[label] at (img22.north west) {Light mask};
    \node[label] at (img23.north west) {OLAT (seed 1)};
    \node[label] at (img31.north west) {Input image};
    \node[label] at (img32.north west) {Light mask};
    \node[label] at (img33.north west) {OLAT (seed 2)};
  \end{tikzpicture}
  \caption{\label{fig:failure_cases}%
    \textbf{Failure cases.}
    \textbf{Top:} the light spread shape is biased towards cones, diverging from the ground truth.
    \textbf{Middle:} decomposition occasionally ignores the provided light mask.
    \textbf{Bottom:} different random seeds can yield different OLAT decompositions.
  }
\vspace{-0.5em}
\end{figure}

%% file: assets/lighting_harmonization_extra.tex
\begin{figure*}[t]
  \centering
  \begin{tikzpicture}
    \node[anchor=south west, inner sep=0] (image) at (0,0) {
        \includegraphics[width=0.97\linewidth]{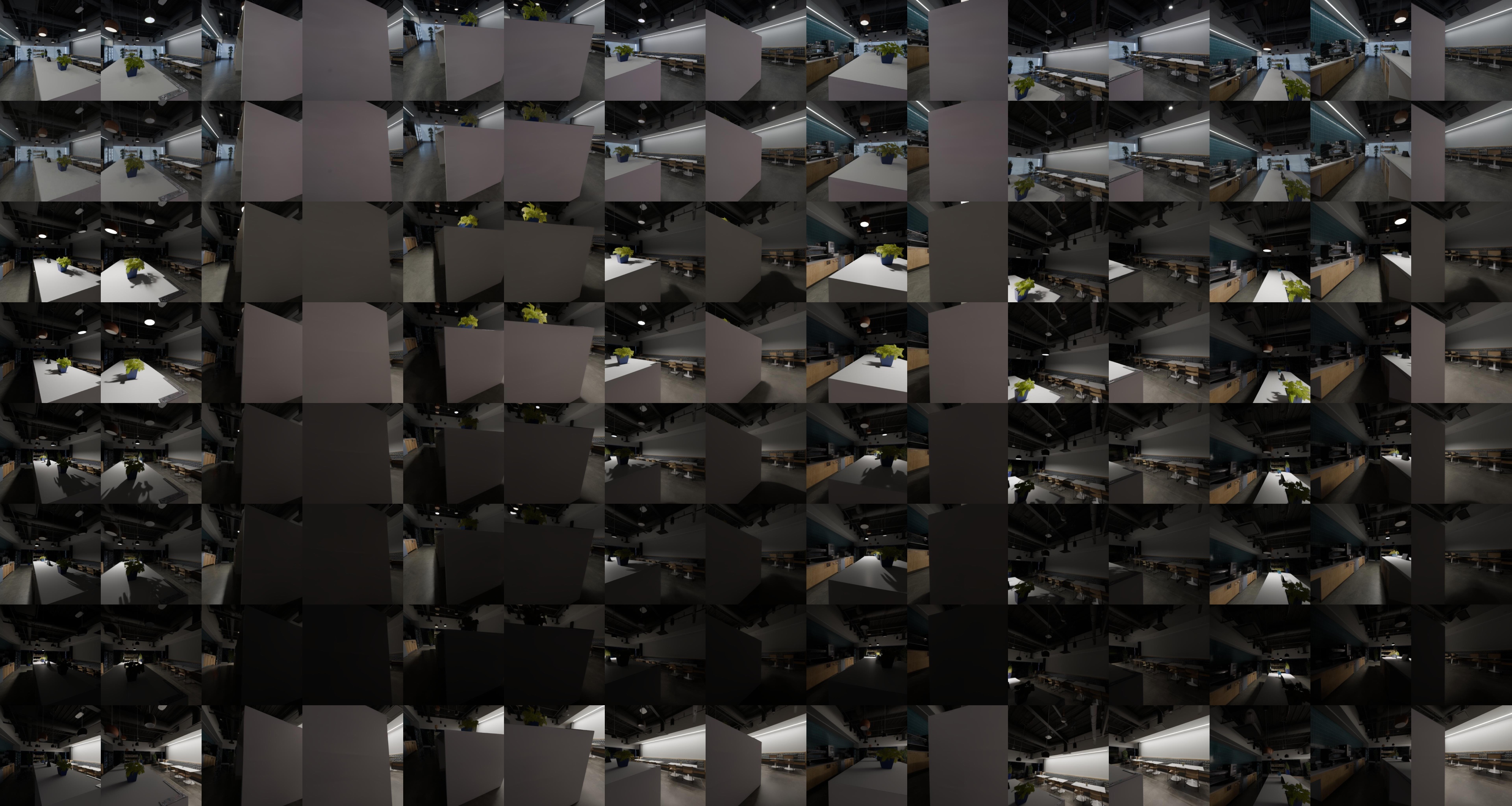}
    };
    \begin{scope}[
        x={($0.1*(image.south east)$)},
        y={($0.1*(image.north west)$)},
        font=\sffamily\footnotesize,
        nodes={text depth=0.25ex,text height=1.25ex}
        ]
        \drawrectangleblue{(0, 0)}{(0.67, 8.75)}
        \node[above,rotate=90] at (0.02, 9.375) {Inputs};
        \node[above,rotate=90] at (0.02, 8.125) {Ambient};
        \node[above,rotate=90] at (0.02, 6.875) {OLAT 1};
        \node[above,rotate=90] at (0.02, 5.625) {OLAT 2};
        \node[above,rotate=90] at (0.02, 4.375) {OLAT 3};
        \node[above,rotate=90] at (0.02, 3.125) {OLAT 4};
        \node[above,rotate=90] at (0.02, 1.875) {OLAT 5};
        \node[above,rotate=90] at (0.02, 0.625) {OLAT 6};
    \end{scope}
  \end{tikzpicture}\\[2pt]
  \begin{tikzpicture}
    \node[anchor=south west, inner sep=0] (image) at (0,0) {
        \includegraphics[width=0.97\linewidth]{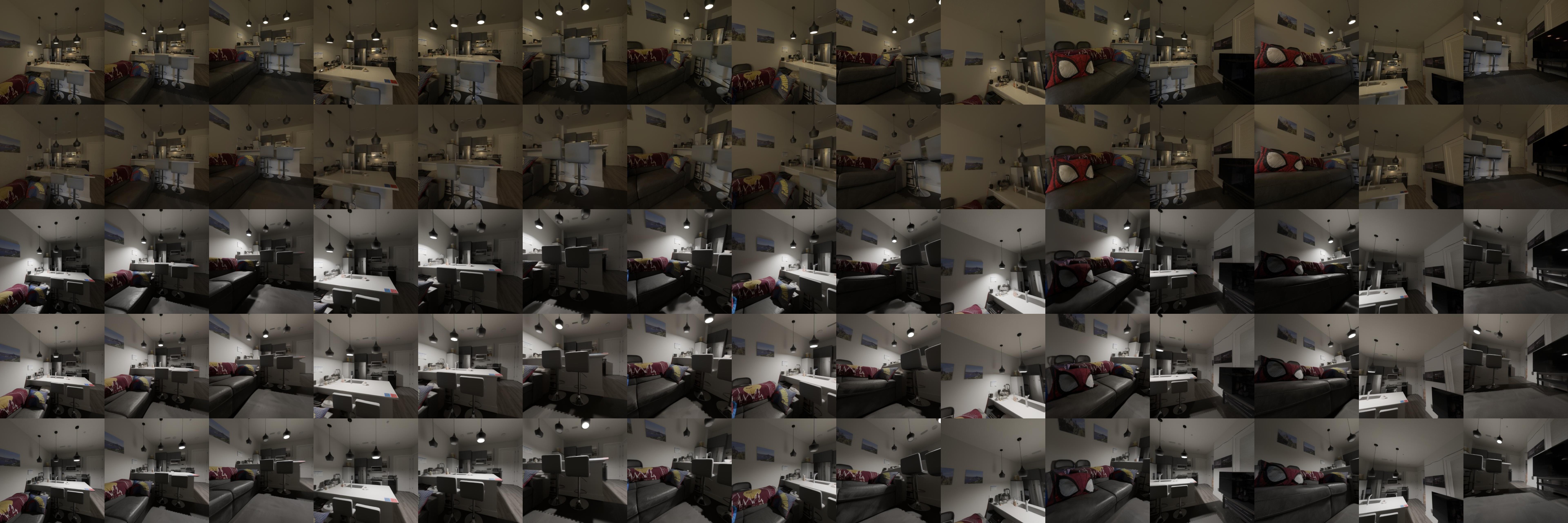}
    };
    \begin{scope}[
        x={($0.1*(image.south east)$)},
        y={($0.1*(image.north west)$)},
        font=\sffamily\footnotesize,
        nodes={text depth=0.25ex,text height=1.25ex}
        ]
        \drawrectangleblue{(0, 0)}{(0.67, 8)}
        \node[above,rotate=90] at (0.02, 9) {Inputs};
        \node[above,rotate=90] at (0.02, 7) {Ambient};
        \node[above,rotate=90] at (0.02, 5) {OLAT 1};
        \node[above,rotate=90] at (0.02, 3) {OLAT 2};
        \node[above,rotate=90] at (0.02, 1) {OLAT 3};
    \end{scope}
  \end{tikzpicture}\\[2pt]
  \begin{tikzpicture}
    \node[anchor=south west, inner sep=0] (image) at (0,0) {
        \includegraphics[width=0.97\linewidth]{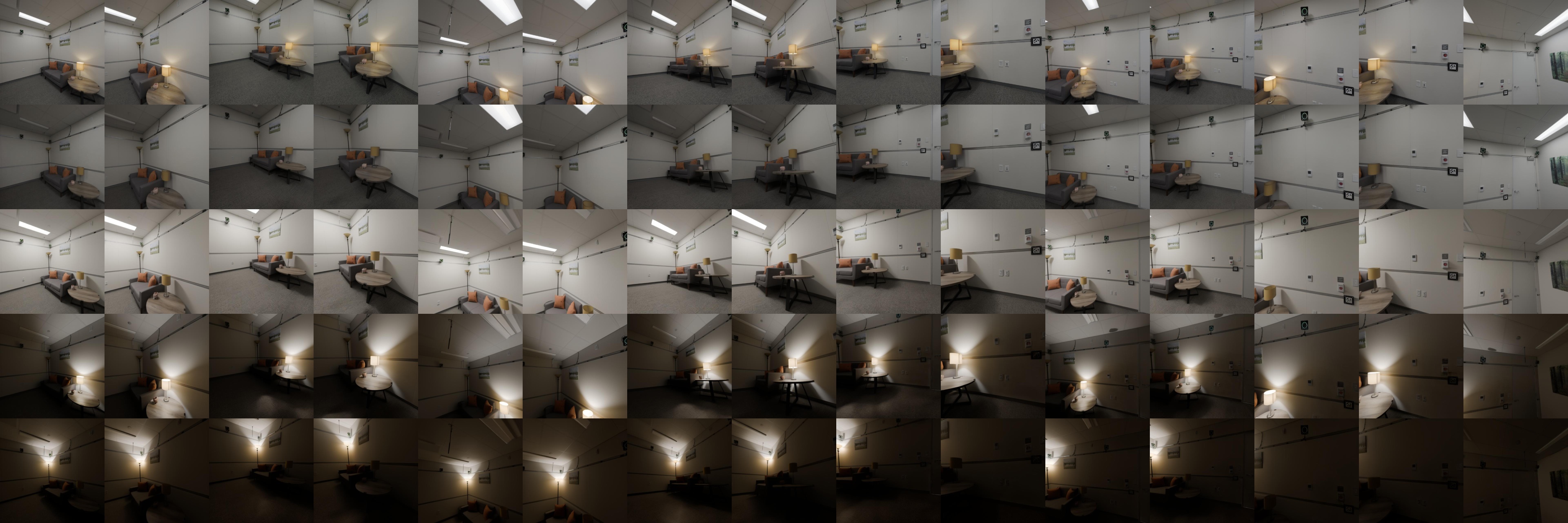}
    };
    \begin{scope}[
        x={($0.1*(image.south east)$)},
        y={($0.1*(image.north west)$)},
        font=\sffamily\footnotesize,
        nodes={text depth=0.25ex,text height=1.25ex}
        ]
        \drawrectangleblue{(0, 0)}{(0.67, 8)}
        \node[above,rotate=90] at (0.02, 9) {Inputs};
        \node[above,rotate=90] at (0.02, 7) {Ambient};
        \node[above,rotate=90] at (0.02, 5) {OLAT 1};
        \node[above,rotate=90] at (0.02, 3) {OLAT 2};
        \node[above,rotate=90] at (0.02, 1) {OLAT 3};
    \end{scope}
  \end{tikzpicture}\\[2pt]
  \caption{\label{fig:lighting_harmonization_extra}%
    \textbf{Additional Multi-view Lighting Harmonization Results.}
    From the first input image, we decompose the lighting into ambient and OLAT components (blue), then propagate them consistently across all views (top row).
    Real-world captures from VR-NeRF \cite{XuALGBKRPKBLZR2023}.
  }
\end{figure*}